\documentclass[preprint,
superscriptaddress,
amsmath,amssymb,aps,showkeys,showpacs,
twoside,final,secnumarabic,nofootinbib]{revtex4-2}
\usepackage[paperwidth=205mm,paperheight=290mm,top=17mm,bottom=25mm,
inner=17mm,outer=17mm,twoside]{geometry}
\usepackage{booktabs}
\usepackage{array}
\usepackage{amssymb, amsmath}
\usepackage[linesnumbered,ruled,vlined,noend]{algorithm2e}
\usepackage{cmap} % Better search in PDF
\usepackage[T1,T2A]{fontenc}
\usepackage[utf8]{inputenc}
\usepackage[russian,english]{babel}
\usepackage{color}
\usepackage{graphicx}% Include figure files
\usepackage{dcolumn}% Align table columns on decimal point
\usepackage{bm} % bold math
\usepackage[unicode=true,colorlinks=true,linkcolor=magenta,urlcolor=blue,citecolor=blue,breaklinks]{hyperref}
\usepackage{amsthm}
\usepackage{mathtools}
\newtheorem{definition}{Definition}
\newtheorem{proposition}{Proposition}
\newtheorem{theorem}{Theorem}[section]
\newtheorem{corollary}{Corollary}[theorem]
\newtheorem{lemma}[theorem]{Lemma}
\newtheorem{remark}{Remark}
\theoremstyle{plain}
\usepackage{pgfplots,tikz}
\usetikzlibrary{shapes,decorations,arrows,positioning,fit}
\pgfplotsset{compat=newest}
\usepackage{grffile}
\usepackage{multirow}
\usepackage{url}
\usepackage{breakurl}
\usepackage{xcolor}
\makeatletter
\DeclareGraphicsExtensions{.eps}
\newcount\issue
\newcount\Vol
\newcount\numb
\usepackage{fancyhdr}
\newtheorem{property}{Property} 
\def\Vol{\textbf{80}}
\def\numb{x}
\begin{document}
%====== Title block ============
\title{Machine Learning in Fundamental Physics \\[20pt]
Natural Image Classification via Quasi-Cyclic Graph Ensembles and Random-Bond Ising Models at the Nishimori Temperature\\} 
\def\addressa{Research and Development Department, T8 LLC, Moscow, 107076 Russia}
\def\addressb{Department of Computer Engineering, South-West State University, Kursk, 305040 Russia}
\author{\firstname{V.S.}~\surname{Usatyuk}}
\email[E-mail: ]{L@Lcrypto.com }
\affiliation{\addressa}
\affiliation{\addressb}
\author{\firstname{D.A.}~\surname{Sapozhnikov}}
\affiliation{\addressa}
\author{\firstname{S.I.}~\surname{Egorov}}
\affiliation{\addressb}
\received{xx.xx.2025}
\revised{xx.xx.2025}
\accepted{xx.xx.2025}
\begin{abstract}
Multi-class image classification with high-dimensional CNN features incurs heavy memory and computation costs, while conventional spectral graph methods fail on complex natural-image manifolds. We introduce a physics-inspired framework that maps frozen MobileNetV2 features to Ising spins on sparse MET-QC-LDPC graphs, forming a Random-Bond Ising Model operated at the Nishimori temperature identified via the Bethe–Hessian spectrum. Our design rests on two cornerstones: a spectral-topological correspondence that links trapping sets to cohomological invariants for suppressing harmful subgraphs, and a fast quadratic-Newton estimator that converges in roughly nine iterations. The method compresses 1280-dimensional features to only 32 dimensions for ImageNet-10 and 64 for ImageNet-100, achieving 98.7\% and 84.92\% top-1 accuracy, respectively, while cutting FLOPs by up to $29\times$ compared with ResNet-50 at a small accuracy cost. The key contributions are a rigorous trapping-set-to-cohomology link, an efficient Nishimori-temperature solver, and proof that topology-guided LDPC embeddings yield highly compressed yet accurate classifiers.
\end{abstract}
\keywords{Clustering, Classification, Topological Coding Theory, Feature Embeddings, Graph codes, Multi-Edge Type graphs, Nishimori, Random Bond Ising Model, Quasi-Cyclic, QC-LDPC, Spectral Gap, Stochastic Block Model \\[5pt]}
\maketitle
\thispagestyle{fancy}
%====== Main text ============
\section{Introduction}\label{sec:intro}
Deploying deep neural networks on bandwidth‑constrained, latency‑sensitive platforms (mobile phones, autonomous vehicles, remote sensors) requires that the classifier head—the linear layer mapping backbone embeddings to class scores—be both memory and communication efficient. When the number of categories $C$ grows to hundreds or thousands (e.g., ImageNet‑100/1000), the head's parameter count ($d\times C$) dominates overall resource consumption.
Many works have attempted to alleviate this bottleneck, reporting the following empirical outcomes:
\begin{itemize}
\item Linear dimensionality reduction (e.g.\ Probabilistic PCA) can shrink the activation dimension before classification, but at the cost of a substantial drop in top‑1 accuracy—up to $41.33\%$ relative loss~\cite{-3,-2}.
\item Low‑rank factorisations of the weight matrix (singular‑value truncation) have yielded modest compression on simple benchmarks such as CIFAR‑10 and STL‑10, yet they do not translate into competitive performance on larger datasets~\cite{-1}.
\item Deep auto‑encoders that learn a non‑linear manifold for the classifier weights achieve higher compression ratios, but their training demands extensive label‑rich data, and the reconstruction objective does not directly optimise classification accuracy; over‑compression often erodes fine‑grained class information~\cite{0,-1}.
\item Cross‑Stage‑Partial (CSP) architectures reduce the backbone embedding size from the typical $d=1280$ (MobileNetV3) to $d=256$, thereby cutting classifier parameters by roughly $88\%$ for ImageNet‑100 and raising top‑1 accuracy from $81.26\%$ to $82.64\%$ after dropout regularization~\cite{00}. Despite this improvement, CSP‑based models still lag behind larger CNNs (ResNet‑50: $86.69\%$) and Vision Transformers ($\approx 86.01\%$) on the same benchmark and lack a flexible mechanism for head ensembling that could trade latency for higher accuracy.
\end{itemize}
Collectively, these approaches either sacrifice predictive performance or impose computational demands incompatible with real‑time, on‑device deployment, leaving an unmet need for a compact yet accurate classifier head.
We address this gap by re‑conceptualising the classification head as a statistical‑physics system. Each class weight vector is treated as a spin on a sparse graph whose edges carry random couplings, forming a Random‑Bond Ising Model (RBIM). By enforcing the Nishimori condition and exploiting Bethe free‑energy approximations, we tune edge weights so that the resulting spin ensemble undergoes a sharp paramagnetic–ferromagnetic transition precisely at the decision boundary. Building on recent findings that quasi‑cyclic low‑density parity‑check (QC‑LDPC) graphs with Nishimori‑weighted edges improve binary clustering of CNN descriptors~\cite{2}, we extend the construction to multi‑class settings.
The key outcomes of this strategy are:
\begin{enumerate}
\item \textit{Extreme compression with competitive accuracy.} The $1\,280\times C$ classifier matrix is compressed to a $32$‑dimensional (for $C=10$) or $64$‑dimensional (for $C=100$) spectral embedding per class. On ImageNet-10 the method attains $98.7\%$ top‑1 accuracy, and on ImageNet-100 it reaches $82.7\%$; a soft ensemble of three QC‑LDPC graphs further lifts performance to $84.9\%$ top‑1, all while requiring only matrix–vector products at inference time.
\item \textit{Spectral–topological correspondence for trapping sets.} We prove that small subgraphs (trapping sets) in the Tanner graph of a QC‑LDPC code generate isolated eigenvalues in both the Bethe‑Hessian and non‑backtracking operators, manifesting as mod‑2 cohomological defects on the classifier manifold. These defects are quantifiable via Stiefel–Whitney classes and Kervaire invariants of the associated chain complex.
\item \textit{Robust Nishimori temperature estimator.} A low‑degree Lagrange interpolant fitted to the smallest eigenvalue of the weighted Bethe–Hessian yields a stable estimate of the critical Nishimori inverse temperature $\beta_{N}$. The algorithm converges reliably even under severe geometric distortion, providing an optimal spin‑glass/paramagnetic threshold for image clustering (unsupervised classification).
\item \textit{Scalable validation.} Experiments on ImageNet-10 and ImageNet-100 confirm that the proposed graph‑based embedding satisfies real‑time constraints without sacrificing accuracy, outperforming prior compression schemes.
\end{enumerate}
In summary, our work introduces a fundamentally different paradigm—leveraging sparse spin‑glass models, spectral graph theory, and topological coding theory—to compress deep classification heads. This yields a highly compact, accurate, and inference‑efficient solution that bridges the performance–resource gap left by existing methods.
The remainder of the paper is organised as follows: Section~\ref{sec:rbim} reviews the RBIM and Nishimori condition; Section~\ref{sec:qcldpc} details QC‑LDPC graph constructions; Appendix~\ref{sec:topology} analyses trapping sets via topological coding theory and information geometry; Section~\ref{sec:nish_est} presents the temperature‑estimation algorithm; Section~\ref{sec:experiments} reports experimental results. Sections~\ref{sec:future} and \ref{sec:conclusion} discuss future directions and conclude.
%%%%%%%%%%%%%%%%%%%%%%%%%%%%%%%%%%%%%%%%%%%%%%%%%%%%%%%%%%%%%
% 3. Random Bond Ising Model
%%%%%%%%%%%%%%%%%%%%%%%%%%%%%%%%%%%%%%%%%%%%%%%%%%%%%%%%%%%%%
\section{Random Bond Ising Models}\label{sec:rbim}
Let $\mathcal G=(V,E)$ be an undirected graph with $|V|=n$.
Assign a binary spin variable $s_i\in\{-1,+1\}$ to each vertex $i\in V$ and denote the
spin configuration by $\mathbf s = (s_1,\dots ,s_n)^{\!\top}$.
For a symmetric coupling matrix $J=(J_{ij})_{i,j=1}^n$, the Hamiltonian of the Random‑Bond Ising Model is
\begin{equation}\label{eq:hamiltonian}
  \mathcal H_J(\mathbf s)= -\sum_{i<j} J_{ij}s_i s_j 
                        = -\tfrac12\,\mathbf s^{\!\top}J\mathbf s .
\end{equation}
At inverse temperature $\beta$ the Boltzmann distribution reads
\begin{equation}\label{eq:boltzmann}
  \mu_{\beta,J}(\mathbf s)= 
     \frac{\exp\bigl(-\beta\,\mathcal H_J(\mathbf s)\bigr)}%
          {Z_{J,\beta}}, \qquad 
  Z_{J,\beta}= \sum_{\mathbf s\in\{-1,+1\}^{n}}\! e^{-\beta\mathcal H_J(\mathbf s)} .
\end{equation}
In a classification setting the coupling $J$ encodes similarity between feature vectors: 
\begin{equation}\label{eq:binary_coupling}
  J_{ij}= 
  \begin{cases}
    +1, & \text{if } y_i=y_j ,\\[2pt]
    -1, & \text{otherwise},
  \end{cases}
\end{equation}
where $y_i$ denotes the class label of vertex $i$.
More refined couplings are obtained from a weighted graph built on the
CNN features (see Section~\ref{sec:graph_construction}).
\begin{remark}[Binary spin–class correspondence]\label{rem:binary}
For binary classification ($C=2$) the Ising variable admits an immediate semantic interpretation:
\begin{equation}\label{eq:spin-class-map}
y_i = 0 \;\Longleftrightarrow\; s_i=-1,\qquad
y_i = 1 \;\Longleftrightarrow\; s_i=+1 .
\end{equation}
Under this identification the rule~\eqref{eq:binary_coupling} is ferromagnetic ($J_{ij}=+1$) for intra‑class pairs and antiferromagnetic ($J_{ij}=-1$) for inter‑class pairs.  Consequently, minimising the Ising Hamiltonian~\eqref{eq:hamiltonian} at low temperature is equivalent to finding a spin configuration that maximises agreement with the ground‑truth binary partition $\{0,1\}$.
\end{remark}
\begin{proposition}[From binary spins to $C$‑ary labels]\label{prop:multiclass-spin}
Let $\mathcal G$ be a graph on $n$ vertices and let $H_{\beta_N,J}$ be the Bethe–Hessian at the Nishimori temperature.  Denote by $\psi^{(1)},\dots,\psi^{(r)}$ the eigenvectors associated with the $r$ smallest \emph{negative} (or near‑null) eigenvalues.  Each eigenvector induces a binary partition
$\mathcal V^{\pm}_\alpha = \{i : \pm\psi^{(\alpha)}_i > 0\}$,
generalising the two‑class Ising problem to an $r$‑bit error‑correcting code.  The concatenated embedding
\begin{equation}\label{eq:multi-embed}
\mathbf e_i = \bigl(\psi^{(1)}_i,\dots,\psi^{(r)}_i\bigr)^{\!\top}\in\mathbb R^r
\end{equation}
provides a unique signature for every class provided $r\ge\lceil\log_2 C\rceil$.  Thus the binary Ising model extends naturally to multi‑class classification by replacing a single discrete spin with an $r$‑dimensional vector of soft spectral spins.
\end{proposition}
%%%%%%%%%%%%%%%%%%%%%%%%%%%%%%%%%%%%%%%%%%%%%%%%%%%%%%%%%%%%%
% 3.1 Nishimori temperature
%%%%%%%%%%%%%%%%%%%%%%%%%%%%%%%%%%%%%%%%%%%%%%%%%%%%%%%%%%%%%
\subsection{Nishimori condition and phase transition}\label{subsec:nishimori}
For a disordered Ising system the \emph{Nishimori line}~\cite{3,4}
specifies a particular temperature $\beta_N$ at which the distribution of couplings
matches that of the spins.  If the edge weights $J_{ij}$ are drawn from a symmetric
distribution $P(J)$ satisfying
\begin{equation}\label{eq:nish_condition}
   P(J) = p(|J|)\,e^{\beta_N J},
   \qquad 
   \int p(|J|)e^{\beta_N J}\,\mathrm d J =1 ,
\end{equation}
the Nishimori condition holds.  At $\beta=\beta_N$ the model exhibits a
paramagnetic–spin‑glass transition, and many observables (e.g.\ magnetisation,
overlap) become analytically tractable.
A convenient characterisation of $\beta_N$ uses the Bethe–Hessian matrix $H_{\beta,J}$
(see Section~\ref{sec:bethe}).  Its smallest eigenvalue $\lambda_{\min}(\beta)$
is \emph{not} monotone: as $\beta$ grows from zero, $\lambda_{\min}$ crosses
zero downward at the ferromagnetic transition $\beta_F$, stays negative, reaches
a minimum, and then rises again.  The Nishimori point is uniquely defined by the
\emph{largest} inverse temperature at which this eigenvalue returns to zero:
\begin{equation}\label{eq:nish_root}
   \lambda_{\min}(\beta_N)=0,\qquad 
   \beta_N = \max\{\beta>0 : \lambda_{\min}(\beta)=0\}.
\end{equation}
Thus estimating $\beta_N$ reduces to locating the second (rightmost) zero of a
one‑dimensional spectral function.
%%%%%%%%%%%%%%%%%%%%%%%%%%%%%%%%%%%%%%%%%%%%%%%%%%%%%%%%%%%%%
% 3.2 Bethe free energy and Hessian
%%%%%%%%%%%%%%%%%%%%%%%%%%%%%%%%%%%%%%%%%%%%%%%%%%%%%%%%%%%%%
\subsection{Bethe free energy and the Bethe--Hessian}\label{sec:bethe}
\begin{figure}[ht]
    \centering
    \includegraphics[width=300pt]{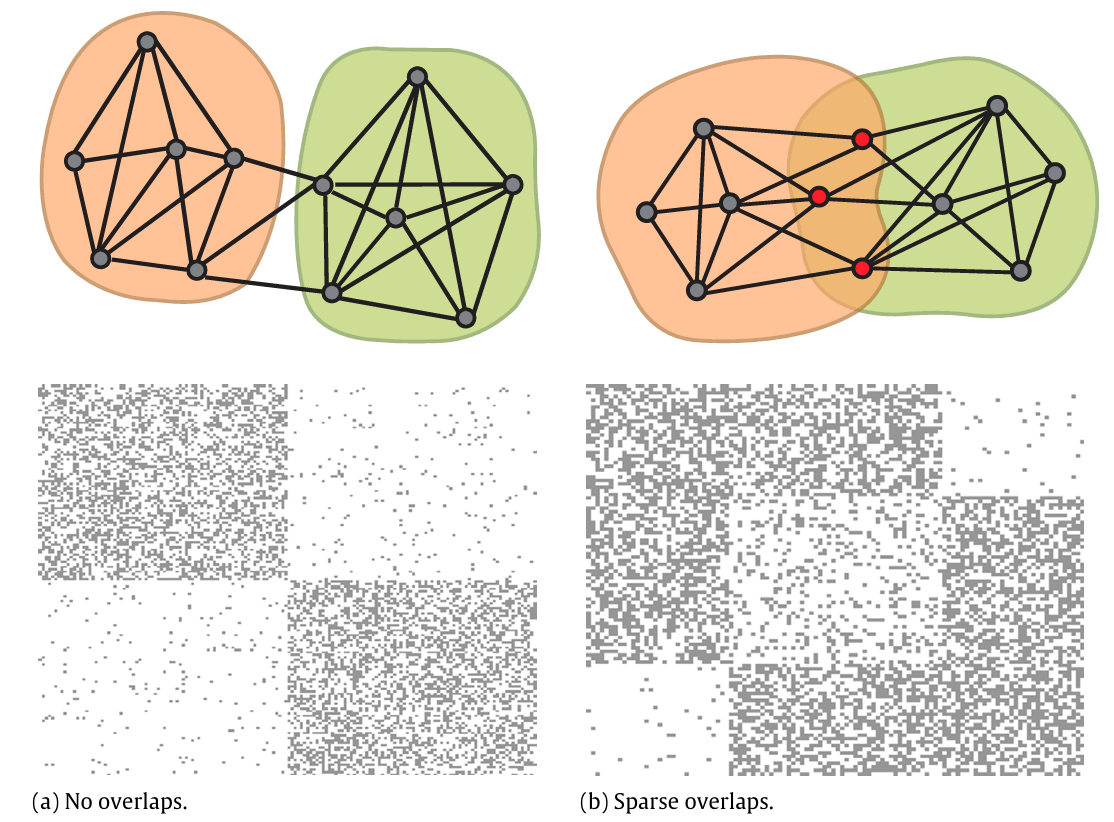}
    \caption{(a) and (b): cluster structures (communities) arising from a phase transition described by adjacency matrices and the associated network graph (cf.\ Fig.~20 in~\cite{5}).}
    \label{Fig_graph_adg}
\end{figure}

Let $q_i=\Pr(s_i=+1)$ be an approximate marginal.
The Bethe free energy associated with a factorised distribution
$p_q(\mathbf s)=\prod_i q_i^{(1+s_i)/2}(1-q_i)^{(1-s_i)/2}$ reads~\cite{6}
\begin{equation}\label{eq:bethe_fe}
  \widetilde F_{J,\beta}(q)
   = \sum_{\mathbf s} p_q(\mathbf s)\bigl[\,
        \beta\,\mathcal H_J(\mathbf s) 
       +\ln p_q(\mathbf s)\bigr].
\end{equation}
The Hessian of $\widetilde F_{J,\beta}$ with respect to the vector
$\mathbf m = (2q_i-1)_i$ is the \emph{Bethe–Hessian} (also called deformed Laplacian)~\cite{7,1}:
\begin{equation}\label{eq:bethe_hessian}
  H_{\beta,J}= 
     \operatorname{diag}\!\Bigl(
       1+\sum_{k\in\partial i}\frac{\tanh^2(\beta J_{ik})}{1-\tanh^2(\beta J_{ik})}
     \Bigr) 
     - \biggl[\frac{\tanh(\beta J_{ij})}{1-\tanh^2(\beta J_{ij})}\biggr]_{i\neq j}.
\end{equation}
At the Nishimori temperature $\beta_N$ the smallest eigenvalue of $H_{\beta,J}$ vanishes,
signalling a critical point where community structure becomes most pronounced, Fig.~\ref{Fig_graph_adg}.
%%%%%%%%%%%%%%%%%%%%%%%%%%%%%%%%%%%%%%%%%%%%%%%%%%%%%%%%%%%%%
% 4. Quasi‑Cyclic LDPC graphs
%%%%%%%%%%%%%%%%%%%%%%%%%%%%%%%%%%%%%%%%%%%%%%%%%%%%%%%%%%%%%
\section{Multi-Edge Type Quasi‑Cyclic LDPC Graphs}\label{sec:qcldpc}
\begin{figure}[ht]
\centering
\includegraphics[width=0.48\textwidth]{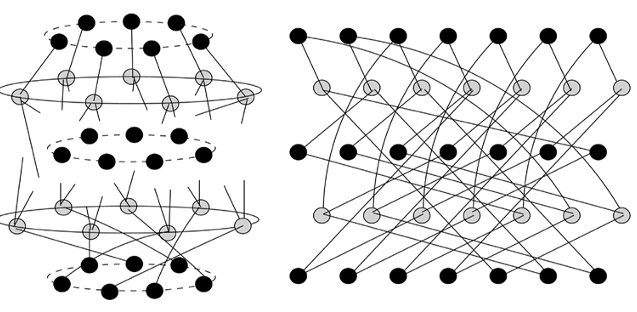}
\includegraphics[width=0.45\textwidth]{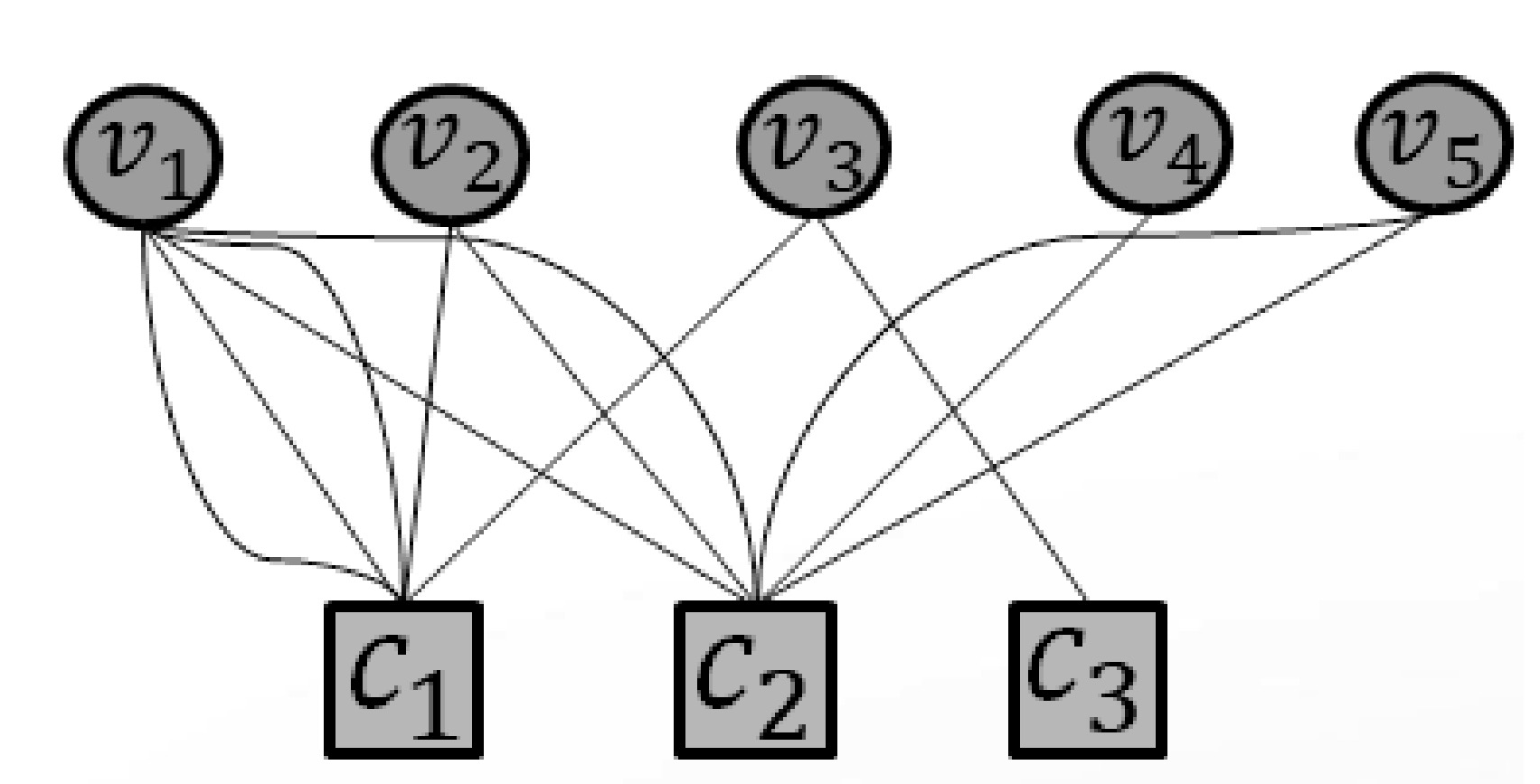}
\caption{(Left) Tanner graph corresponding to the parity-check matrix $H$, composed of $2$ and $3$ rings of size $7$ (cf.\ Fig.~2 in~\cite{11}). (Right) Multi-graph protograph $M(H_{2})$ representation of the QC-LDPC code $H_{2}$.}
\label{proto}
\end{figure}

Low-density parity-check (LDPC) codes naturally define sparse bipartite graphs.
Quasi-Cyclic LDPC (QC-LDPC) codes provide a structured and hardware-friendly subclass defined by a quasi-cyclic parity-check matrix $H$~\cite{8}. An $(N, K)$ QC-LDPC code consists of $N$ total codeword bits, with $K$ information bits and $N-K$ parity bits. The associated Tanner graph is described by a parity-check matrix $H \in \mathbb{F}_2^{mL \times nL}$, constructed from square blocks of size $L \times L$, where each block is either a zero matrix or a circulant permutation matrix (CPM)~\cite{9,10}.
A CPM $P \in \{0,1\}^{L \times L}$ is defined as
\[
P_{ij} = 
\begin{cases}
1, & \text{if } i + 1 \equiv j \pmod{L}, \\
0, & \text{otherwise}.
\end{cases}
\]
Let $P_k$ denote a circulant permutation matrix corresponding to a right shift of the identity matrix $I$ by $k \in \{0, 1, \dots, L-1\}$, i.e.\ the ring $\mathbb{Z}/L\mathbb{Z}$. Then a general parity-check matrix $H_{\text{QC}}$ takes the block form
\[
H_{\text{QC}} = 
\begin{bmatrix}
P_{a_{11}} & P_{a_{12}} & \dots & P_{a_{1n}} \\[2pt]
P_{a_{21}} & P_{a_{22}} & \dots & P_{a_{2n}} \\[2pt]
\vdots & \vdots & \ddots & \vdots \\[2pt]
P_{a_{m1}} & P_{a_{m2}} & \dots & P_{a_{mn}}
\end{bmatrix},
\]
where each $a_{ij} \in \mathcal{A}_L = \{0, 1, \dots, L-1\}$. The circulant size $L$ controls both the code length and the degree of quasi-cyclicity.
When the parity-check matrix consists of quasi-cyclic rings of circulants, we refer to this configuration as a \emph{toroidal graph family}. 
As an example of a toroidal graph, consider a QC-LDPC code defined by the parity-check matrix~\cite{11}
\[
H = 
\begin{bmatrix}
I_1 & I_2 & I_4 \\[2pt]
I_6 & I_5 & I_3
\end{bmatrix},
\]
of size $31 \times 21$ with circulant size $L=7$, employing a two-ring ($3$ columns and $2$ rows) construction $(\mathbb{Z}/7\mathbb{Z})$ as illustrated in Fig.~\ref{proto} (left).
From $H_{\text{QC}}$, two key matrices are derived: the exponent matrix $E(H)$, containing shift values $a_{ij}$, and the protograph matrix $M(H_{\text{QC}})$, where each nonzero CPM is replaced by $1$ and zeros are left as is.
An illustrative example of a protograph is given in Fig.~\ref{proto}. Consider
\[
H_{2} = 
\begin{bmatrix}
I_1 + I_2 + I_7 & I_9 & I_{23} & 0 & 0 \\[2pt]
I_{12} + I_{37} & I_{19} & 0 & I_{32} & I_{11} + I_{12} \\[2pt]
0 & 0 & I_{33} & 0 & 0
\end{bmatrix},
\]
where $I_k$ denotes a CPM shifted by $k$. The use of CPM sums characterises a Multi-Edge Type (MET) QC-LDPC code~\cite{12,13}, allowing more flexible degree distributions and improved performance under iterative decoding. A family of MET QC-LDPC graph codes built from two independent circulant rings will be called \emph{toroidal graphs}.
In the special case where the parity-check matrix contains only one circulant ring with multiple weights, we speak of a \emph{spherical graph family}. For example, such a matrix with CPM size $L=2600$ is~\cite{14,15}
\begin{equation}\label{eq:H3}
\begin{multlined}
H_3 = 
\bigl[\;
I_{2}+I_{3}+I_{5}+I_{280}+I_{437}+I_{511}+I_{636}+I_{797} \\
+I_{1022}+I_{1093}+I_{1233}+I_{1671}+I_{1718}+I_{2254}+I_{2334}
\;\bigr].
\end{multlined}
\end{equation}

The graph-theoretic structure of these codes induces cycles in the corresponding Tanner graph. A block-cycle of length $2l$ corresponds to a sequence of CPMs $\{P_{a_1}, P_{a_2}, \dots, P_{a_{2l}}\}$ satisfying the cycle consistency condition~\cite{10}
\[
\sum_{i=1}^{2l} (-1)^i a_i \equiv 0 \pmod{L}.
\]
Appendix~\ref{sec:topology} presents the required cycle properties and examines how topological invariants affect spectral embedding.
The adjacency matrix of the resulting bipartite graph is
\begin{equation}\label{eq:adjacency_bipartite}
 A=
   \begin{bmatrix}
      0 & H_{\text{QC}}^{\!\top} \\[2pt]
      H_{\text{QC}} & 0
   \end{bmatrix},
\qquad 
D=\operatorname{diag}(A\mathbf 1).
\end{equation}
This adjacency serves as the basis for the coupling matrix $J$ in the RBIM.
%%%%%%%%%%%%%%%%%%%%%%%%%%%%%%%%%%%%%%%%%%%%%%%%%%%%%%%%%%%%%
% 4.1 Nishimori estimator
%%%%%%%%%%%%%%%%%%%%%%%%%%%%%%%%%%%%%%%%%%%%%%%%%%%%%%%%%%%%%
\subsection{Quadratic--Newton estimation of the Nishimori temperature $\beta_N$}\label{sec:nish_est}
The root condition~\eqref{eq:nish_root} can be solved efficiently because, once $\beta>\beta_{SG}$,
the map $\beta\mapsto \lambda_{\min}(\beta)$ is smooth and ultimately increasing after its minimum.
Algorithm~\ref{alg:nishimori} describes the proposed $\beta_N$ estimator. The algorithm typically converges within $\sim 9$ Arnoldi calls, a $6$-fold reduction compared with classical bisection~\cite{1,2}.

\begin{algorithm}[H]
  \caption{Quadratic–Newton estimation of the Nishimori temperature $\beta_N$ on weighted graphs}
  \label{alg:nishimori}
  \SetKwInOut{Require}{Require}
  \Require{Weighted adjacency $J$, lower bound $\beta_{\ell}\approx\hat\beta_{SG}$,
           tolerance $\varepsilon$.}
  $t\gets0$\;
  Initialize interval with $\beta_1=\beta_{\ell}$, 
    $\beta_u\leftarrow c\cdot\max|J_{ij}|$ (or any safe upper bound)\;
  \Repeat{\normalfont convergence}{
    Choose three trial points 
      $(\beta_1,\beta_2,\beta_3)$ in $[\beta_\ell,\beta_u]$
      with $\beta_2=(\beta_\ell+\beta_u)/2$\;
    \For{$k=1,2,3$}{%
       Build $H_{\beta_k,J}$ via Eq.~\eqref{eq:bethe_hessian}\;
       Evaluate $\lambda_k=\lambda_{\min}(H_{\beta_k,J})$ via Arnoldi\;
    }
    Fit a quadratic polynomial 
      $p(\beta)=a\beta^2+b\beta+c$
      to $(\beta_k,\lambda_k)_{k=1}^3$\;
    Compute the positive root  
      $\displaystyle\tilde\beta=
        \frac{-b-\sqrt{b^{2}-4ac}}{2a}$\;
    \eIf{$|\lambda_{\min}(H_{\tilde\beta,J})|<\varepsilon$}{%
      \Return $\beta_N:=\tilde\beta$\;
    }{
      Perform one Newton correction using the Rayleigh quotient:
      let $\mathbf x$ be the eigenvector for $\lambda_{\min}(H_{\tilde\beta,J})$,
      define $f(\beta')=\mathbf x^{\!\top}H_{\beta',J}\mathbf x$, solve
      $f(\beta_{t+1})=0$ for $\beta_{t+1}> \tilde\beta$\;
    }
    $t\gets t+1$\;
  }
\end{algorithm}

\begin{figure}[h]
    \centering
    \includegraphics[width=0.8\linewidth]{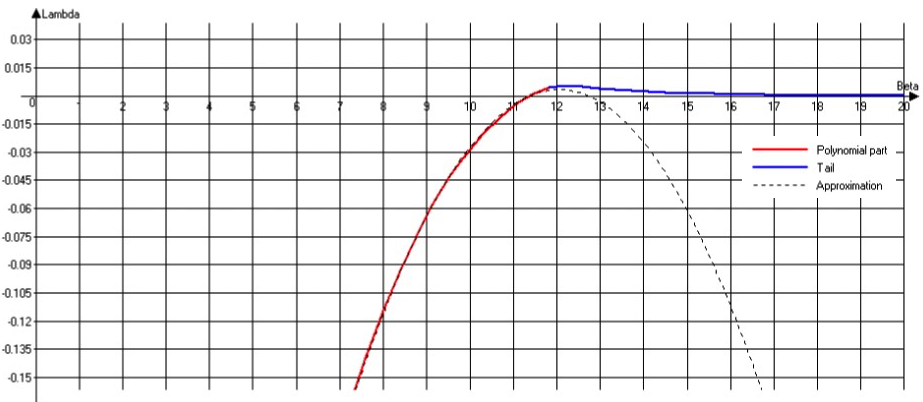}
    \caption{
        Dependence of the smallest eigenvalue $\lambda_{\min}$ on the temperature parameter $\beta$.
        The red curve shows the polynomial part, the blue curve the tail,
        and the black dashed line is a quadratic approximation to the polynomial part.
        The coefficient of determination for the fit is $R^{2}=0.9998$.}
    \label{fig:nishimori}
\end{figure}

%%%%%%%%%%%%%%%%%%%%%%%%%%%%%%%%%%%%%%%%%%%%%%%%%%%%%%%%%%%%%
% 4.2 Graph construction
%%%%%%%%%%%%%%%%%%%%%%%%%%%%%%%%%%%%%%%%%%%%%%%%%%%%%%%%%%%%%
\subsection{Graph construction from CNN features}\label{sec:graph_construction}
Given an image $\mathbf x$, the backbone network $f_{\theta}$ (MobileNetV2, \cite{34}) yields a feature vector 
$\mathbf h\in\mathbb R^{1280}$.
For each class $c$ we retain a subset $\mathcal I^{(c)}$ of $s\ll 1280$
most discriminative indices, e.g.\ those with largest absolute difference
between class‑conditional means.
The reduced feature vector for class $c$ is $\mathbf z^{(c)}
   = (h_{i})_{i\in\mathcal I^{(c)}}\in\mathbb R^{s}$. 
A similarity kernel builds a weighted adjacency matrix
between the $K$ class representatives (cf.\ Eq.~8 in~\cite{2}):
\begin{equation}\label{eq:similarity}
A_{cd}= 
  \begin{cases}
     \exp\!\bigl(-\,\gamma\, d_{\mathrm{cos}}(\mathbf z^{(c)},\mathbf z^{(d)})^{2}\bigr), & c\neq d, \\[4pt]
     0, & c=d,
  \end{cases}
\end{equation}
where $\gamma$ is the kernel bandwidth and $d_{\mathrm{cos}}$ the cosine distance. To enforce sparsity we retain only the $p$ strongest edges per vertex.
The resulting sparse adjacency is then mapped onto a MET QC‑LDPC parity‑check matrix
by selecting an appropriate protograph and CPMs (Section~\ref{sec:qcldpc}).
%%%%%%%%%%%%%%%%%%%%%%%%%%%%%%%%%%%%%%%%%%%%%%%%%%%%%%%%%%%%%
% 5. Experiments
%%%%%%%%%%%%%%%%%%%%%%%%%%%%%%%%%%%%%%%%%%%%%%%%%%%%%%%%%%%%%
\section{Experimental evaluation}\label{sec:experiments}
\begin{figure}[ht]
    \centering
    \includegraphics[width=0.75\linewidth]{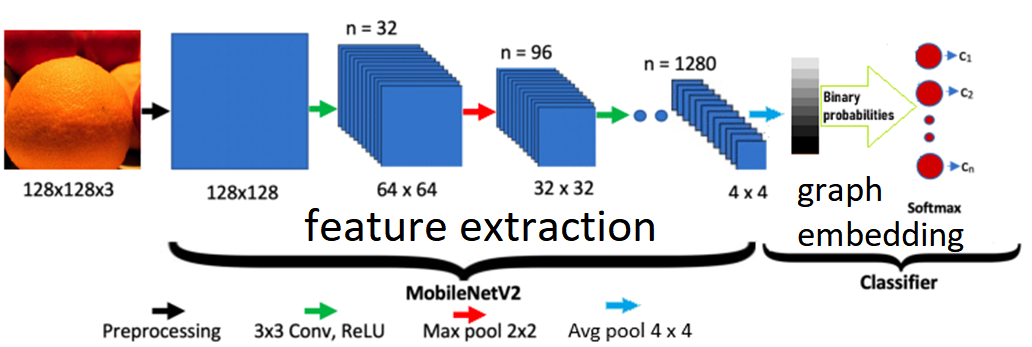}
    \caption{Pipeline: MobileNetV2 feature extraction followed by graph spectral embedding.}
    \label{fig:MobileNetV2}
\end{figure}

All experiments use the same MobileNetV2 backbone pre‑trained on ImageNet, Fig.~\ref{fig:MobileNetV2}.
The feature extractor (CNN layers) is frozen; only the graph embedding (which replaces the feed-forward MLP) and final classifier are trained. We use the following datasets:
\begin{itemize}
\item \textbf{ImageNet-10}: 10‑class subset ($13\,000$ images, $10\,000$ training, $3\,000$ testing).
\item \textbf{ImageNet-100}: 100‑class subset ($130\,000$ training, $5\,000$ testing; $1\,300$ training and $50$ test samples per class).
\end{itemize}
ImageNet-10 uses a 32-D spectral embedding and ImageNet-100 a 64-D embedding.
\begin{figure}[h]
     \centering 
    \includegraphics[width=240pt]{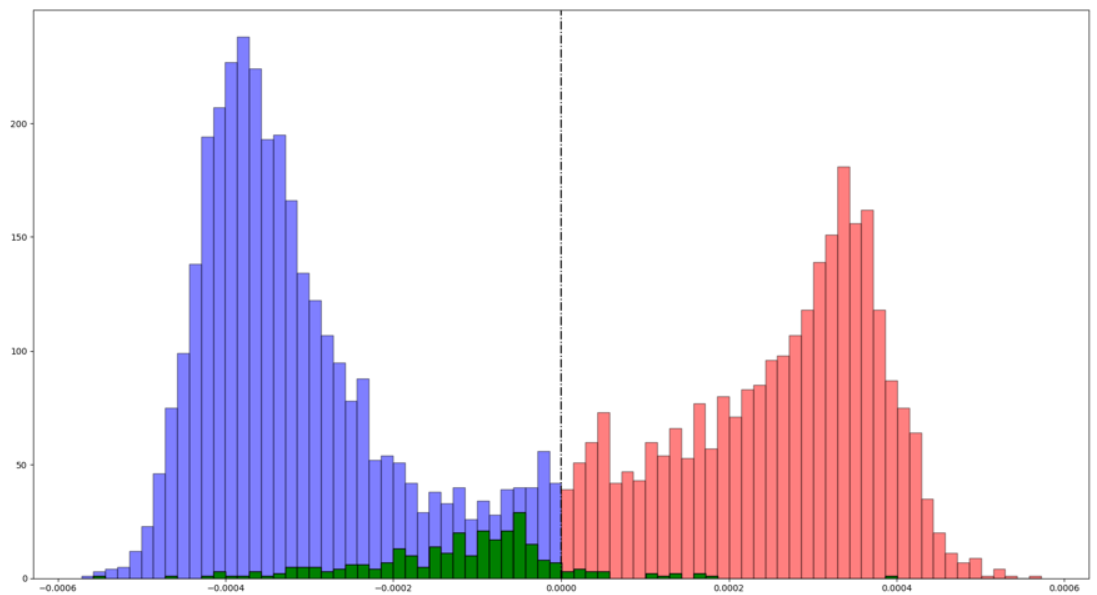}
    \includegraphics[width=240pt]{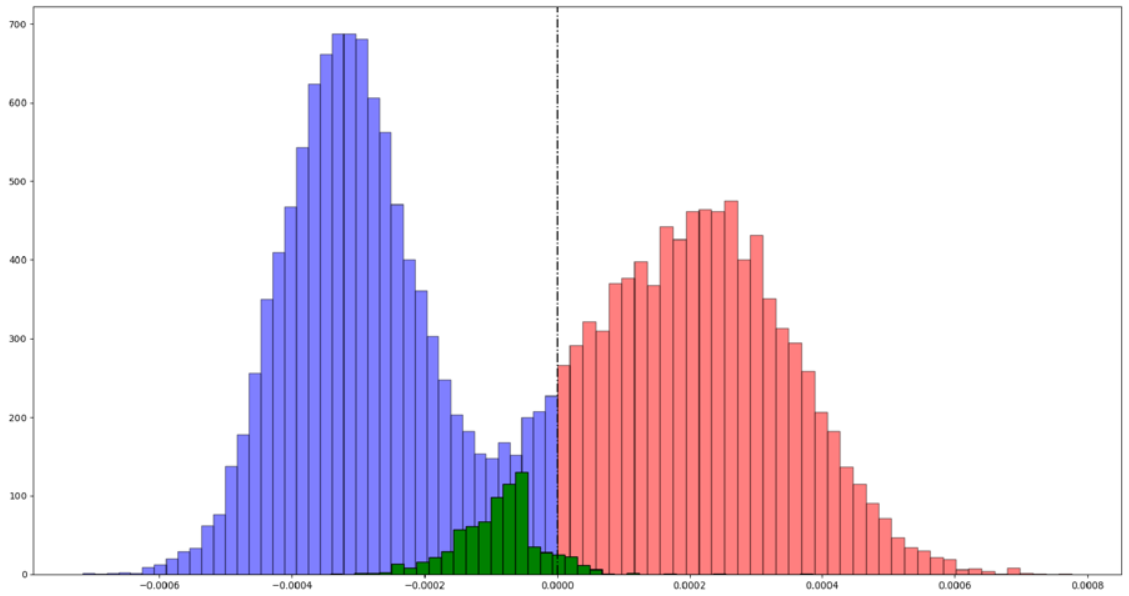}\\
     \includegraphics[width=230pt]{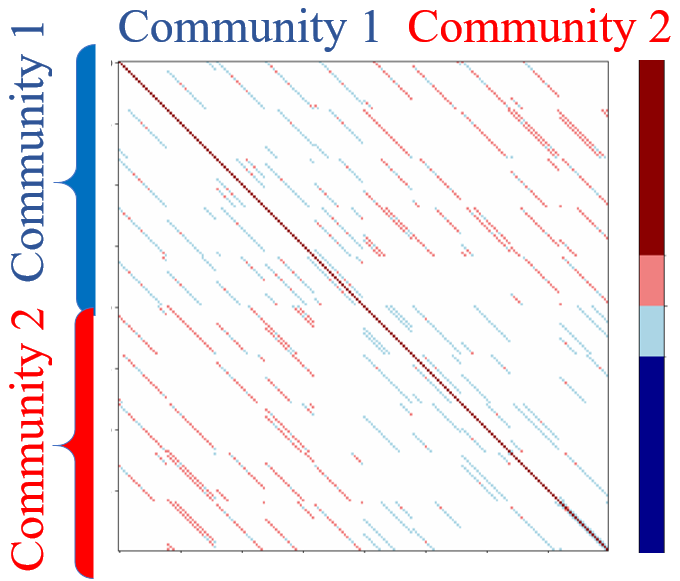}
    \includegraphics[width=230pt]{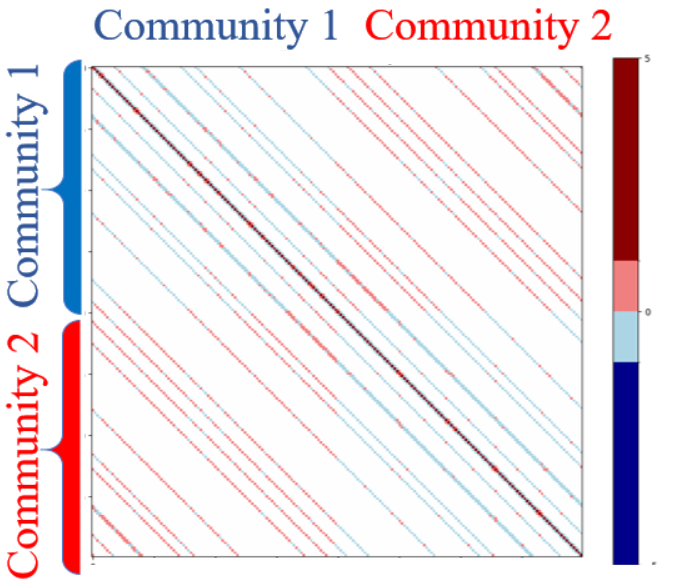}
    \caption{Spectral embedding histogram with adjacency matrix of clusters (visualised by colours) under the estimated Nishimori temperature $\beta_N$. (Left) MET QC-LDPC toroidal graph. (Right) QC spherical graph.}
    \label{fig:SBMs_Hist}
\end{figure}
%%%%%%%%%%%%%%%%%%%%%%%%%%%%%%%%%%%%%%%%%%%%%%%%%%%%%%%%%%%%%
% 5.1 Graph ensembles
%%%%%%%%%%%%%%%%%%%%%%%%%%%%%%%%%%%%%%%%%%%%%%%%%%%%%%%%%%%%%
\subsection{Graph Ensembles for ImageNet‑100}\label{sec:graph-ensembles}
We construct two families of MET–QC‑LDPC graphs and then combine them into an ensemble that produces a final class decision by majority voting, Fig.~\ref{fig:SBMs_Hist}.
\begin{itemize}
\item \textit{Spherical graphs} – each instance consists of a single circulant ring, has fixed column weight $k=10$ and CPM size $L=2600$.
\item \textit{Toroidal graphs} – each instance is built from two independent circulant rings (forming a torus), uses $L=520$, and exhibits an average column weight of $3.6$.
\end{itemize}
Both families are first optimised with respect to the permanent bound~\eqref{eq:perm_bound} and the Bethe‑permanent bound~\cite{27}.  This optimisation explicitly suppresses low‑weight trapping sets that would otherwise degrade the spectral embedding.
From the two families we generate three affinity graphs (see Fig.~\ref{fig:SBMs_Hist}),
feed each graph into an identical classifier and obtain three independent predictions
$\widehat y_1,\;\widehat y_2,\;\widehat y_3$.
The final label is obtained by a simple majority vote.  In the rare case that the
three votes do not agree we invoke an \emph{arbiter network} to resolve the conflict.
Formally, the hard ensemble reads
\begin{equation}\label{eq:arbiter}
\widehat y_{\text{res}}=
\begin{cases}
\widehat y_1, & 
\text{if } \widehat y_1=\widehat y_2=\widehat y_3 , \\[4pt]
\operatorname{Arb}\!\bigl(\widehat y_1,\widehat y_2,\widehat y_3\bigr), &
\text{otherwise},
\end{cases}
\end{equation}
where $\operatorname{Arb}(\cdot)$ denotes the output of the arbiter network.
Instead of hard voting we can average the class‑wise posterior probabilities supplied
by each graph,
\[
p_c = \frac{1}{3}\sum_{i=1}^{3} p^{(i)}_c, \qquad 
\widehat y_{\text{soft}} = \arg\max_{c} p_c ,
\]
with $p^{(i)}_c$ the probability of class $c$ estimated by the $i$‑th graph.
The soft‑voting rule often yields smoother confidence scores while preserving the
benefits of ensemble diversity.
\begin{table}[h]
    \centering
    \caption{Classification results on ImageNet-10 using graph embeddings.}
    \label{tab:results}
    \begin{tabular}{|c|c|c|c|c|}
        \hline
        Class & Precision & Recall & F1-Score & Support \\
        \hline
        0 & 1.00 & 0.98 & 0.99 & 300 \\
        1 & 0.99 & 0.99 & 0.99 & 300 \\
        2 & 1.00 & 0.99 & 0.99 & 300 \\
        3 & 0.99 & 0.94 & 0.96 & 300 \\
        4 & 0.97 & 0.97 & 0.97 & 300 \\
        5 & 0.99 & 0.99 & 0.99 & 300 \\
        6 & 0.94 & 0.95 & 0.95 & 300 \\
        7 & 0.98 & 0.98 & 0.98 & 300 \\
        8 & 0.94 & 0.98 & 0.96 & 300 \\
        9 & 0.96 & 1.00 & 0.98 & 300 \\
        \hline
        \multicolumn{5}{|c|}{\textbf{Aggregate Metrics}} \\
        \hline
        Accuracy & — & — & 0.98 & 3000 \\
        Macro Avg & 0.98 & 0.98 & 0.98 & 3000 \\
        Weighted Avg & 0.98 & 0.98 & 0.98 & 3000 \\
        \hline
    \end{tabular}
\end{table}
%%%%%%%%%%%%%%%%%%%%%%%%%%%%%%%%%%%%%%%%%%%%%%%%%%%%%%%%%%%%%
% 5.2 Training protocol
%%%%%%%%%%%%%%%%%%%%%%%%%%%%%%%%%%%%%%%%%%%%%%%%%%%%%%%%%%%%%
\subsection{Training protocol}
For each class we compute $\mathbf z^{(c)}$, build a sparse similarity graph,
embed it via the Bethe–Hessian at the estimated Nishimori temperature
$\beta_N$ (Algorithm~\ref{alg:nishimori}), and extract the $r$
eigenvectors ($r=32$ for ImageNet-10, $r=64$ for ImageNet-100), Fig.~\ref{fig:SBMs_Hist}.
These eigenvectors constitute a low‑dimensional embedding $\mathbf e_i$ for every training sample.
A linear classifier (softmax) is trained on $\{\mathbf e_i\}$ using cross‑entropy loss.
\begin{figure}
  \centering
  \includegraphics[width=0.48\textwidth]{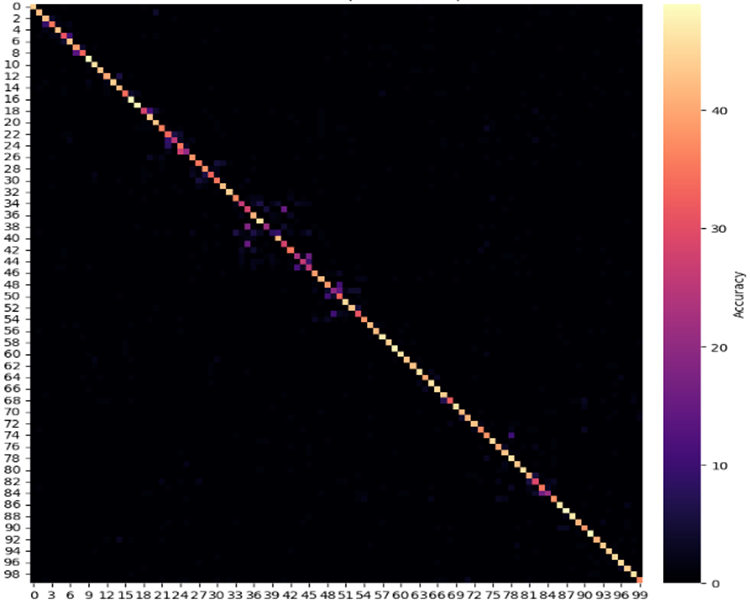}
  \includegraphics[width=0.48\textwidth]{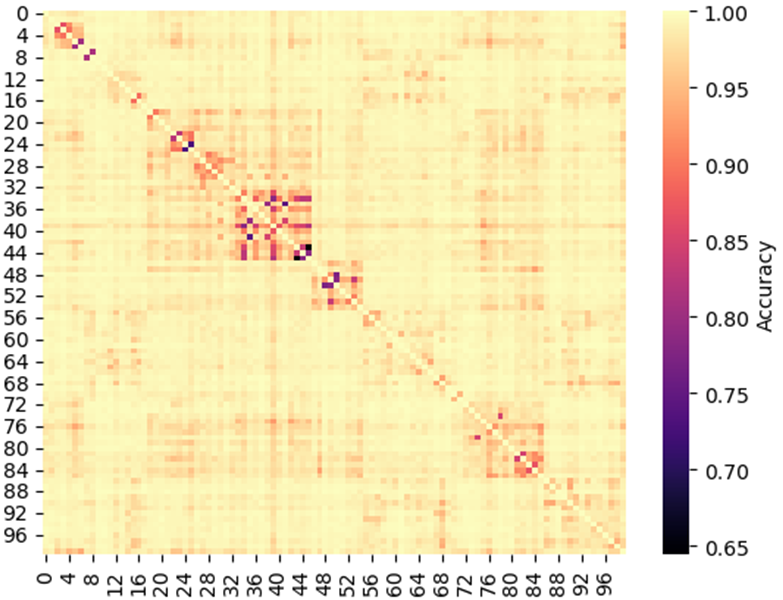}
  \caption{(Left) Confusion matrix heatmap for ImageNet-100.
           (Right) Per‑class top‑1 accuracy heatmap.}
  \label{fig:confusion100}
\end{figure}
\begin{figure}
    \centering
    \includegraphics[width=0.7\linewidth]{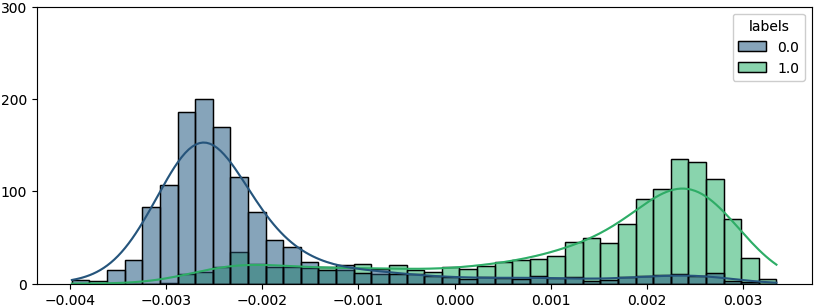}
    \caption{Spectral embedding histogram for the `cock' and `hen' classes.}
    \label{fig:Most_problem}
\end{figure}
%%%%%%%%%%%%%%%%%%%%%%%%%%%%%%%%%%%%%%%%%%%%%%%%%%%%%%%%%%%%%
% 5.3 Results
%%%%%%%%%%%%%%%%%%%%%%%%%%%%%%%%%%%%%%%%%%%%%%%%%%%%%%%%%%%%%
\subsection{Results and Discussion}
Table~\ref{tab:results} reports top-1 accuracy, precision, recall and $F_1$ scores for ImageNet-10. Spherical graph spectral embedding achieves $98.7\%$ top-1 accuracy on ImageNet-10, while Fig.~\ref{fig:confusion100} shows the confusion matrix for ImageNet-100. The mixed ensemble combining spherical and toroidal graphs through majority voting yields the best performance.
For ten classes, discriminative features rarely overlap, but scaling to one hundred classes causes many informative dimensions to intersect, reducing accuracy for both plain CNNs and spectral classifiers. Our experiments identified optimal graphs for ImageNet-100 consisting of a spherical adjacency with weight $k=10$ and circulant size $L=2600$, paired with a toroidal graph having weight $k=10$, protograph size $26 \times 26$ (containing five non-zero MET CPMs of weight $2$), and CPM size $L=100$, as visualised in Fig.~\ref{fig:SBMs_Hist}. This configuration delivers $77.84\%$ top-1 accuracy when processed through the spectral pipeline.
To address frequent confusion between the two highest class probabilities, we introduced a lightweight MLP arbitrator trained on the same embeddings. The arbitrator achieves $98.43\%$ average pairwise accuracy on training data, though the most challenging pairs remain at approximately $67\%$ accuracy. During inference, the system examines the three largest spectral scores $\{p_{(1)}, p_{(2)}, p_{(3)}\}$, invoking the arbitrator when the margin $p_{(1)} - p_{(2)}$ falls below a preset threshold $T$. This uncertainty-driven fallback mechanism boosts top-1 accuracy from $77.84\%$ to $\mathbf{82.73\%}$ while maintaining the original runtime efficiency and memory footprint of the single-graph approach.
The Erdős–Rényi graph spectral embedding demonstrates $69.1\% \pm 2.88\times10^{-5}$ precision under Nishimori temperature across five trials, averaged over all random seeds. Our three-graph hard ensemble achieves $\mathbf{82.7\%}$ accuracy on ImageNet-100 while using just one twentieth of the original feature dimension and memory footprint, representing a $13.6\%$ improvement over standard Erdős–Rényi graphs without large weights.
Classification difficulties are most evident between class 7 (``cock'') and class 8 (``hen''), as shown in Fig.~\ref{fig:confusion100}. Although the network reaches a training accuracy of $85\%$ on these two categories, validation performance collapses to $63\%$, indicating severe over‑fitting. The spectral embedding histogram obtained at the Nishimori temperature (Fig.~\ref{fig:Most_problem}) reveals that the corresponding Gaussian clusters intersect with heavy tails in this problematic region, which explains the pronounced confusion.
A second troublesome area occurs around class 44 (``spider''). In the heat-map visualisation of stochastic block‑model embeddings (Fig.~\ref{fig:SBMs_Hist}), a three-class overlap can be observed, suggesting that the underlying feature distributions are not well separated.
To alleviate these ambiguities, we construct a soft ensemble consisting of three $64$‑dimensional graph embeddings. The three vectors are concatenated and fed to a classifier trained with standard softmax cross‑entropy loss. This strategy directly addresses the heavy‑tailed intersections observed in Fig.~\ref{fig:Most_problem} by smoothing decision boundaries through complementary relational information.
The empirical impact of the proposed approach on ImageNet‑100 is reported in Table~\ref{tab:dnn-comparison}, cf.~\cite{00}. Compared with our hard‑embedding baseline, the soft ensemble raises Top‑1 accuracy from $82.70\%$ to $84.92\%$, while only modestly increasing model size (from $1.09\,\text{M}$ to $1.33\,\text{M}$ parameters) and computational cost (from $117.1\,\text{M}$ to $141.3\,\text{M}$ FLOPs). Although the soft‑embedding model is still considerably smaller than heavyweight architectures such as ResNet‑50 ($86.01\%$, $23.7$M parameters) and Vision Transformer ($88.69\%$, $85.7$M parameters), it attains a competitive accuracy–efficiency trade‑off.
\begin{table}[h]
    \centering
    \caption{Top‑1 accuracy, number of parameters and FLOPs for various DNN models~\cite{00}.}
    \label{tab:dnn-comparison}
    \begin{tabular}{l c c c}
        \toprule
        Model & ImageNet-100 Top‑1 Accuracy (\%) & Params & FLOPs \\
        \midrule
        Vgg16                     & 82.07 & 134.7M & 15.5G   \\
        MobileNetV2               & 82.60 & 2.4M   & 313.0M  \\
        MobileNetV3               & 83.15 & 4.3M   & 224.9M  \\
        CSPdarknet53              & 75.50 & 1.1M   & 159.5M  \\
        LightCSPNet dropout (p=0.2)  & 82.64 & 0.9M   & 143.9M  \\
        Hard Ensemble graph embedding (our) & 82.70 & 1.09M  & 117.1M  \\
        Soft Ensemble graph embedding (our) & 84.92 & 1.33M  & 141.3M  \\
        ResNet50                  & 86.01 & 23.7M  & 4109.7M \\
        Vision Transformer        & 88.69 & 85.7M  & 16.9G   \\
        \bottomrule
    \end{tabular}
\end{table}
%%%%%%%%%%%%%%%%%%%%%%%%%%%%%%%%%%%%%%%%%%%%%%%%%%%%%%%%%%%%%
% 6. Future directions
%%%%%%%%%%%%%%%%%%%%%%%%%%%%%%%%%%%%%%%%%%%%%%%%%%%%%%%%%%%%%
\section{Future Directions}\label{sec:future}
The findings of this paper suggest several concrete research avenues. First, the present pipeline freezes the MobileNetV2 backbone, uses a cosine similarity kernel, and adopts a fixed MET‑QC‑LDPC adjacency; an obvious extension is to make all three components differentiable and optimise them jointly. One could learn the feature extractor $f_{\theta}$ together with the graph embedding, parameterise the similarity kernel (for instance by learning a Mahalanobis metric) and back‑propagate through the adjacency construction, and finally adjust the circulant shift values that define the QC‑LDPC parity‑check matrix by gradient‑based or reinforcement‑learning methods. Rich multimodal embeddings such as those produced by Contrastive Language-Image Pre-training (CLIP) could serve as powerful initialisations for this end‑to-end training.
A second line of inquiry is to embed topological and information‑geometric penalties directly into the loss function. Since trapping sets correspond to non‑trivial mod‑2 cohomology and curvature of the belief manifold—manifested as near‑zero eigenvalues of the Bethe–Hessian—it is natural to add differentiable proxies for these obstructions. Possible regularisers include a term proportional to the sum of negative Bethe–Hessian modes (penalizing flat directions), an estimator of the Wu class $w_2$ via cycle parity, or curvature corrections to the Bethe free energy obtained from the Amari–Chentsov tensor. By penalising these quantities during training we could suppress harmful subgraphs dynamically rather than removing them in a post‑hoc design step.
Scaling the methodology to larger and more diverse data domains is another promising direction. Extending experiments from ImageNet‑10/100 to the full ImageNet-1k benchmark will test whether the observed compression ratios and accuracy gains persist at million‑scale sample sizes. Moreover, applying the framework to video streams—where temporal coherence can be encoded as an additional edge type—and to multi-modal inputs (e.g., image–text pairs) by constructing heterogeneous MET‑QC‑LDPC graphs could reveal new ways of handling high‑dimensional, structured data while retaining a small memory footprint suitable for edge devices.
The sparse, highly structured adjacency matrices produced by our graph design are also attractive as priors for modern Graph Neural Networks. Incorporating them into message‑passing GCN or GAT layers, using them to sparsify self‑attention mechanisms in Transformers, or employing diffusion‑based graph convolutions may combine the interpretability of physics‑inspired graphs with the expressive power of learned GNNs. An intriguing sub‑question is whether the Nishimori temperature $\beta_{N}$ can be treated as a learnable hyperparameter within such architectures.
Finally, several theoretical questions remain open. It would be valuable to derive rigorous bounds linking reductions in Stiefel–Whitney obstructions (or other cohomological defects) to improvements in classification margins; to explore connections between Ihara–Bass zeta function poles and the higher cup products recently studied in quantum LDPC codes~\cite{li2024}; and to generalise the quadratic‑Newton Nishimori estimator to Potts or vector spin models that might better capture multi‑class structure. Addressing these issues will deepen our understanding of how statistical‑physics and topological coding concepts translate into practical machine‑learning tools.
%%%%%%%%%%%%%%%%%%%%%%%%%%%%%%%%%%%%%%%%%%%%%%%%%%%%%%%%%%%%%
% 7. Conclusions
%%%%%%%%%%%%%%%%%%%%%%%%%%%%%%%%%%%%%%%%%%%%%%%%%%%%%%%%%%%%%
\section{Conclusions}\label{sec:conclusion}
We introduced a physics‑inspired embedding that maps high‑dimensional CNN descriptors onto spins of a Random‑Bond Ising Model defined on specially constructed MET‑QC‑LDPC graphs. The central theoretical insight is a spectral–topological correspondence: trapping sets in the Tanner graph generate poles of the Ihara–Bass zeta function, which appear as isolated eigenvalues of both the non‑backtracking operator and the Bethe–Hessian. These spectral signatures encode mod‑2 cohomology classes—Stiefel–Whitney obstructions, Kervaire invariants and curvature defects on the belief manifold.
On the algorithmic side we proposed a quadratic‑Newton estimator for the Nishimori temperature operating directly on the weighted Bethe–Hessian. By evaluating the smallest eigenvalue at three trial temperatures above $\beta_{SG}$, fitting a quadratic interpolant and applying a Rayleigh‑quotient Newton correction, convergence is reached in roughly nine Arnoldi iterations, yielding about a sixfold speed‑up over conventional bisection while preserving numerical accuracy.
The graph design pipeline combines permanent and Bethe‑permanent bounds with ACE/EMD optimisation to suppress low‑genus trapping sets, producing two families of sparse graphs (spherical and toroidal). These structures compress the $1280$‑dimensional MobileNetV2 features to $32$ dimensions for a ten‑class ImageNet subset and to $64$ dimensions for a hundred‑class subset. Despite this compression we obtain top‑1 accuracies of \textbf{98.7\%} on ImageNet‑10 and \textbf{84.9\%} on ImageNet‑100, while reducing FLOPs by up to $29\times$. Thus, by unifying statistical mechanics, topological coding theory, information geometry and algebraic graph design, the proposed method delivers compact yet accurate classifiers that are well suited for energy‑constrained edge deployment.

\appendix
%%%%%%%%%%%%%%%%%%%%%%%%%%%%%%%%%%%%%%%%%%%%%%%%%%%%%%%%%%%%%
% A. Topological Coding Theory & Statistical Manifolds
%%%%%%%%%%%%%%%%%%%%%%%%%%%%%%%%%%%%%%%%%%%%%%%%%%%%%%%%%%%%%
\section{Topological Coding Theory and Spectral Signatures of Trapping Sets}\label{sec:topology}
This appendix develops two complementary viewpoints on LDPC subgraphs.  The first (Sections~\ref{subsec:chain}–\ref{subsec:ktheory}) treats the Tanner graph as a chain complex and studies its mod‑2 cohomological invariants—Stiefel–Whitney classes, the Kervaire invariant, cobordism, and $K$‑theory—following the framework of topological coding theory~\cite{li2024}.  The second (Sections~\ref{subsec:statmanifold}–\ref{subsec:loops}) interprets Belief Propagation as optimisation on a statistical manifold, showing that trapping sets are simultaneously curvature defects and cohomological obstructions. We then examine how these topological signatures manifest spectrally (Section~\ref{subsec:spectral}), via the Ihara--Bass zeta function (Section~\ref{subsec:ihara}), through permanent bounds (Section~\ref{subsec:perm}), and in downstream classification performance (Section~\ref{sec:TS_influence}).

%-----------------------------------------------------------
% A.1 Chain Complex
%-----------------------------------------------------------
\subsection{Chain complex of an LDPC code}\label{subsec:chain}
Let $H\in\mathbb F_2^{m\times n}$ be the parity‑check matrix of a (quasi-cyclic) LDPC code and let $\mathcal G=(V\cup C,E)$ be its Tanner graph, where $V$ indexes variable nodes and $C$ indexes check nodes.  We view $\mathcal G$ as the 1‑skeleton of a cell complex $X_H$ whose 2‑cells are attached along trapping‑set cycles (or more generally along closed walks in the factor graph).  The cellular chain complex over $\mathbb F_2$ is
\[
C_2(X_H;\mathbb F_2)\xrightarrow{\partial_2}C_1(X_H;\mathbb F_2)
\xrightarrow{\partial_1}C_0(X_H;\mathbb F_2),
\]
where $\partial_1$ is the edge‑vertex incidence and $\partial_2$ encodes the attaching maps of 2‑cells.  Because $H$ itself defines the check‑to‑variable adjacency, the transpose $H^{\!\top}$ can be identified with $\partial_2$ (up to orientation).  The first homology group
\[
H_1(X_H;\mathbb F_2)=\ker\partial_1/\operatorname{im}\partial_2
\]
counts independent non‑bounding cycles; its dimension is the first Betti number $\beta_1$.  A trapping set $TS(a,b)$ induces a subgraph whose own $\beta_1>0$ whenever it contains at least one independent cycle.

%-----------------------------------------------------------
% A.2 Stiefel–Whitney classes
%-----------------------------------------------------------
\subsection{Stiefel--Whitney classes and local orientability}\label{subsec:sw}
For a finite CW complex $X$, the total Stiefel--Whitney class $w(X)=1+w_1+w_2+\cdots$ lives in $H^*(X;\mathbb Z/2)$.  The class $w_1\in H^1(X;\mathbb Z/2)$ is the obstruction to orientability of the tangent bundle (for manifolds) or, more generally, to lifting the structure group of the underlying real vector bundle through $\mathrm{SO}\to\mathrm O$.  
In the LDPC setting we consider the flat line bundle over $X_H$ determined by the edge signs of the Ising coupling matrix $J$: each edge carries weight $J_{ij}$ and parallel transport around a cycle acquires factor $\prod_{e\in C}\operatorname{sgn}(J_e)\in\{\pm1\}$.  Reducing mod 2, this defines a class
\[
w_1(\mathcal T)=[\operatorname{sgn}(J)]\in H^1(\mathcal T;\mathbb Z/2),
\]
for any trapping set $\mathcal T$.  The following property is immediate:
\begin{property}
A trapping set subgraph $\mathcal T$ has trivial $w_1$ iff every simple cycle in $\mathcal T$ contains an even number of antiferromagnetic edges (i.e.\ the gauge transformation of Lemma~\ref{lem:gauge_app} is globally consistent).  
\end{property}
When $w_1(\mathcal T)\neq 0$, no global ferromagnetic gauge exists; BP message passing experiences frustration analogous to a non‑orientable surface.  
The second class $w_2\in H^2(X_H;\mathbb Z/2)$ appears once we attach 2‑cells (checks) to the 1‑skeleton.  It measures whether the induced local frames on the 1‑skeleton extend over the 2‑cells.  In LDPC terms, a non‑trivial $w_2$ means that the parity constraints cannot all be satisfied simultaneously by any local orientation of the variable bits—precisely the situation in an undetected trapping set with odd syndrome ($b>0$).  Consequently:
\begin{itemize}
\item $\mathcal T$ is \emph{orientable} ($w_1=0$) iff it supports a consistent BP gauge;
\item $w_2(\mathcal T)\neq 0$ signals that the local check equations are globally incompatible, producing pseudocodewords.
\end{itemize}

%-----------------------------------------------------------
% A.3 Kervaire invariant
%-----------------------------------------------------------
\subsection{Kervaire invariant and quadratic forms on codes}\label{subsec:kervaire}
Let $\mathcal T$ be a trapping set with $w_1(\mathcal T)=0$.  On its first cohomology group $V=H^1(\mathcal T;\mathbb Z/2)$ the cup product defines a symmetric bilinear form
\[
B(x,y)=\langle x\cup y,[\mathcal T]\rangle\in\mathbb Z/2 .
\]
Because $\mathcal T$ is bipartite (as a subgraph of an LDPC Tanner graph) we have $x\cup x=0$, so $B$ is symplectic.  A \emph{quadratic refinement} of $B$ is a function $q:V\to\mathbb Z/2$ such that
\[
q(x+y)-q(x)-q(y)=B(x,y).
\]
Such refinements exist because $w_1=0$, and the Arf invariant
\[
\kappa(\mathcal T)=\sum_{x\in V}(-1)^{q(x)}\in\{0,\pm2^{|V|/2}\},
\]
normalized modulo 8, is the \emph{Kervaire invariant}.  In topological coding theory it distinguishes two fundamental classes of trapping sets:
\begin{itemize}
\item $\kappa=0$: the quadratic form is even; pseudocodewords supported on $\mathcal T$ have effective weight in $4\mathbb Z$.
\item $\kappa\neq 0$: the form is odd; the trapping set supports logical operators (or pseudocodewords) of half‑integer ``Kervaire parity'', making them robust against local BP updates.
\end{itemize}
For the small trapping sets reported in Table~\ref{tab:ts-comparison}, the Kervaire column records exactly this Arf invariant computed on $H^1(\mathcal T;\mathbb Z/2)$.

%-----------------------------------------------------------
% A.4 Cobordism
%-----------------------------------------------------------
\subsection{Cobordism, bordism triviality, and topological obstructions}\label{subsec:cobordism}
Two closed complexes $X_0$ and $X_1$ are \emph{(unoriented) cobordant} if there exists a compact $(d+1)$‑complex $W$ with $\partial W=X_0\sqcup X_1$.  In the LDPC context, a code deformation is precisely such a cobordism: we transform one protograph into another by a sequence of local edge surgeries (adding/removing checks and variables) that trace out a higher‑dimensional complex $W$.
The Stiefel--Whitney numbers $w_I[X]$ are complete invariants of unoriented bordism.  Hence, if two LDPC subgraphs differ in any S--W number, no local surgery path (cobordism) can connect them without passing through a singular intermediate code—typically a phase transition where the BP decoder fails because the spectral gap collapses.
In the present work we use this obstruction as a \emph{design rule}: ACE/EMD optimisation is restricted to surgeries that preserve all S--W numbers of the protograph.  Equivalently, we only allow cobordisms inside the same bordism class; trapping sets belonging to a different class cannot be removed by any sequence of local edge flips compatible with the quasi‑cyclic lift.  This matches the inductive ``interlaced family'' construction of Li et al.~\cite{li2024}, where homological invariants are preserved when lifting small seed codes to infinite families.

The same tools apply to the feature manifolds arising from CNN embeddings. Let $G=(V,E)$ be a simple graph and let $\phi:V\rightarrow\mathbb R^{d}$ embed its vertices as points of a smooth manifold 
$M_{G}:=\phi(V)\subset\mathbb R^{d}$~\cite{20}.
We call $G$ \emph{bordism‑trivial} if $M_{G}$ is null–bordant in the appropriate bordism group,
i.e.\ there exists a compact $(d+1)$‑manifold $W$ with
\[
\partial W = M_{G}\qquad(\text{or } \partial W=M_{G}\sqcup(-M_{G}) 
\text{ for the unoriented case}),
\]
and all characteristic numbers of $M_{G}$ vanish:
\[
w_{k}(M_{G})=0,\; k>0,
  \qquad
  p_{j}(M_{G})=0,\; j\ge1.
\]
Equivalently $[M_{G}]=0$ in the oriented bordism group $\Omega^{SO}_{d}$ 
(and in $\Omega^{O}_{d}$ when orientation is ignored).
For a multi‑class problem let $\{M_{i}\}_{i=1}^{K}$ be the manifolds obtained from
the feature vectors of each class.
A \emph{bordism obstruction} between classes $i$ and $j$
occurs if the disjoint union 
\[
M_{ij}:= M_{i}\sqcup(-M_{j})
\]
is not null‑bordant in $\Omega^{SO}_{d}$ (or $\Omega^{O}_{d}$).  
A convenient detection criterion is the non‑vanishing of a characteristic number,
e.g.
\[
\exists\,k>0:\; w_{k}[M_{ij}]\neq0
  \quad\text{or}\quad
  \exists\,j:\; p_{j}[M_{ij}]\neq0.
\]
If such an obstruction exists for any pair of classes, no graph‑based embedding can achieve perfect separation in the high‑dimensional feature space~\cite{21}.

%-----------------------------------------------------------
% A.5 Higher signatures and K-theory
%-----------------------------------------------------------
\subsection{$K$--theory, higher signatures and the continuous genus}\label{subsec:ktheory}
Belief Propagation propagates messages $m_{i\to j}\in\Delta^1$ (the 1‑simplex of binary probabilities).  The collection $\mathcal B=\{(m_{i\to j})\}_{(ij)\in E}$ over all directed edges forms a section of a vector bundle $E_{\rm BP}$ whose base is the directed edge set and whose fiber is $\mathbb R^2$ (logits).  At a BP fixed point, the messages are covariantly constant with respect to the local constraint factors.
The formal difference $[E_{\rm BP}]-[\underline{\mathbb R}^{|E|}]$ defines an element of the reduced $K$‑theory $\widetilde K^0(X_H)$. 

Given a cell complex $\mathcal M_{G}$ derived from the graph,
let $L_{k}=d_{k-1}^{*}d_{k-1}+d_{k}d_{k}^{*}$ be the combinatorial Hodge Laplacian acting on $k$‑cochains~\cite{22}.
The kernel dimension satisfies 
\[
\dim\ker L_{k}= \beta_{k},
\]
so that spectral information directly yields Betti numbers.
For any cohomology class $x\in H^{*}(B\pi_{1}(\mathcal M_{G});\mathbb Q)$,
the higher signature 
\[
\sigma_{x}(\mathcal M_{G})
   = \bigl\langle L(\mathcal M_{G})\smile x,\,[\mathcal M_{G}] \bigr\rangle
\]
pairs the Hirzebruch $L$‑class with the fundamental class~\cite{23}.
By the Novikov conjecture, $\sigma_{x}\!\pmod{2}$ detects non‑trivial torsion,
hence non-bounding trapping cycles render some higher signature non-zero.

The Dirac operator on the bipartite graph,
\[
\mathcal D=
\begin{pmatrix}
0 & A_{\!vn} \\
A_{\!vn}^{\!\top} & 0
\end{pmatrix},
\]
has spectrum symmetric about zero~\cite{24,25}.
Its analytic index equals the difference of dimensions of positive and negative eigenspaces,
which, by the Atiyah–Singer index theorem, coincides with $\beta_{0}-\beta_1$~\cite{26}.
Thus a non-zero index signals the presence of topological defects (e.g.\ non-contractible cycles).
In $K$‑theory, the Kasparov matrix
\[
D_{\mathrm{Kas}}=
\begin{pmatrix}
0 & S & T \\
S^{\!\top} & 0 & 0 \\
T^{\!\top} & 0 & 0
\end{pmatrix},
\]
built from suitable incidence matrices $S$ and $T$, satisfies:
$K_{0}= \dim\ker D_{\mathrm{Kas}}^{2}$ (counts zero‑modes, i.e.\ $\beta_{0}$);
$K_{1}= \operatorname{rank}D_{\mathrm{Kas}}\bmod 2$ (detects $\mathbb Z/2$ torsion, equivalent to the Kervaire invariant)~\cite{27}.
Non‑trivial $K$‑groups therefore provide a concise algebraic signature of the topological obstruction introduced by a trapping set.

The \emph{continuous genus} is a scalar number that captures how ``twisted'' or ``curved'' a graph $G$ is, by looking at the spectrum of a matrix derived from the graph:
\[
\widehat A(H)=
\frac{1}{2\sqrt{n_V}}
\Biggl(
   \sum_{\lambda_i\in\Lambda^{+}}\!\sqrt{\lambda_i}
   -
   \sum_{\lambda_j\in\Lambda^{-}}\!\sqrt{-\lambda_j}
\Biggr),
\]
where $n_V$ is the number of vertices in the graph.
The experimental study of how topological invariants influence spectral embedding is presented in Section~\ref{sec:TS_influence}.

%-----------------------------------------------------------
% A.6 Statistical manifold of BP
%-----------------------------------------------------------
\subsection{The statistical manifold and Bethe free energy}\label{subsec:statmanifold}
BP can be seen as optimisation on a \emph{statistical manifold} $\mathcal M$~\cite{yedidia2001}.  Points on $\mathcal M$ are collections of marginal distributions $b_i(x_i)$ (beliefs at variable nodes) and factor beliefs $b_a(\mathbf x_{\partial a})$ at check nodes.  The natural divergence is the Kullback–Leibler divergence, not Euclidean distance.
For an arbitrary trial distribution $b\in\mathcal M$, the variational Bethe free energy is
\[
F_{\rm Bethe}(b)=\sum_{a}\sum_{\mathbf x_{\partial a}}b_a(\mathbf x_{\partial a})
\ln\frac{b_a(\mathbf x_{\partial a})}{\psi_a(\mathbf x_{\partial a})}
-\sum_i(d_i-1)\sum_{x_i}b_i(x_i)\ln b_i(x_i),
\]
where $\psi_a$ are the factor potentials derived from the Ising couplings.
Yedidia, Freeman and Weiss proved that \emph{BP fixed points are exactly the stationary points of $F_{\rm Bethe}$ on $\mathcal M$}.  The Hessian of $F_{\rm Bethe}$ evaluated at the paramagnetic point is precisely the Bethe–Hessian matrix $H_{\beta,J}$ of Eq.~\eqref{eq:bethe_hessian}.

%-----------------------------------------------------------
% A.7 I-projections and flatness
%-----------------------------------------------------------
\subsection{I--projections and information geometry}\label{subsec:iproj}
Information geometry provides a generalized Pythagorean theorem for $\mathcal M$.  The manifold decomposes into an \emph{e‑flat} submanifold of exponential families (consistent marginals) and an \emph{m‑flat} submanifold of mixture families (moment matching).  BP performs alternating I‑projections: message updates project a current belief onto the submanifold defined by one parity check, then back to the product manifold of independent variables.
For a tree‐like graph these manifolds intersect orthogonally and BP converges in one sweep—the geometry is flat.  When loops are present (as in trapping sets) the Amari–Chentsov curvature tensor
\[
R_{ijk\ell}=\partial_i\Gamma_{jk}^\ell-\partial_j\Gamma_{ik}^\ell+
\Gamma_{im}^\ell\Gamma_{jk}^m-\Gamma_{jm}^\ell\Gamma_{ik}^m
\]
does not vanish.  The curvature is localized exactly on the independent cycles counted by $\beta_1$.  Consequently BP cannot reach the true maximum a posteriori (MAP) solution; it converges instead to a local minimum of $F_{\rm Bethe}$—a pseudocodeword.

%-----------------------------------------------------------
% A.8 Loops, curvature and trapping sets
%-----------------------------------------------------------
\subsection{Trapping sets as joint topological and curvature defects}\label{subsec:loops}
A trapping set $\mathcal T$ manifests simultaneously in three languages:
\begin{itemize}
\item \textbf{Cohomological} (Sec.~\ref{subsec:sw}--\ref{subsec:kervaire}): non‑trivial $H^1(\mathcal T;\mathbb Z/2)$, possibly carrying non‑zero S--W or Kervaire obstructions.
\item \textbf{Spectral} (Appendix~\ref{app:traps}): each independent cycle contributes one near‑zero mode of the Bethe–Hessian at the Nishimori temperature.
\item \textbf{Information geometric}: the same cycles produce curvature in $\mathcal M$, preventing flat I‑projections and trapping BP in a wide basin.
\end{itemize}
Table~\ref{tab:topology} summarizes the unified dictionary.
\begin{table}[htbp]
\centering
\caption{Dictionary of topological coding theory for LDPC subgraphs.}
\label{tab:topology}
\setlength{\tabcolsep}{4pt}
\resizebox{\textwidth}{!}{%
\begin{tabular}{lll}
\toprule
\textbf{Topological tool} & \textbf{LDPC / BP counterpart} & \textbf{Physical meaning} \\
\midrule
Stiefel--Whitney $w_1,w_2$ &
Parity-check constraints as obstruction to orientability &
Local consistency vs.\ global ability to gauge all edges ferromagnetically. \\[2pt]
Kervaire invariant $\kappa$ &
Quadratic form on $H^1(\mathcal T;\mathbb{Z}/2\mathbb{Z})$ from cup product &
Determines parity (even/odd) of pseudocodeword weights; robustness to local perturbations. \\[2pt]
Cobordism $[X]=0\in\Omega_*^{O}$ &
Code deformation path between two protographs &
Whether one code can be reached from another by ACE/EMD surgery without closing the spectral gap. \\[2pt]
$K$--theory / Dirac index &
Bundle of BP messages over the Tanner graph &
Classification of stable vs.\ unstable decoding states; analytic index $=\beta_0-\beta_1$. \\
\bottomrule
\end{tabular}%
}
\end{table}
Because curvature is localized on cycles, removing a trapping set (by ACE/EMD optimization or protograph surgery that changes its bordism class) simultaneously:
\begin{enumerate}
\item lowers $\beta_1$ and removes an $H^1$ generator,
\item eliminates one near‑zero Bethe–Hessian mode,
\item flattens the corresponding direction in $\mathcal M$, allowing BP to converge closer to the true marginals.
\end{enumerate}
This is why trapping-set suppression yields both better LDPC decoding and, by analogy, cleaner spectral embeddings for classification.

%-----------------------------------------------------------
% A.9 Spectral manifestation
%-----------------------------------------------------------
\subsection{Spectral manifestation of trapping sets}\label{subsec:spectral}
Trapping sets (TS) in Tanner graphs correspond to local topological defects that disrupt the decoding dynamics of LDPC codes~\cite{16}. A trapping set $TS(a, b)$, formed by cycles (block-cycles for QC-LDPC) or cycle overlaps, consists of $a$ variable nodes and $b$ odd-degree check nodes, where the configuration prevents successful iterative decoding. When $b = 0$, such sets correspond to codewords, i.e.\ $TS(a, 0)$ with $a = d_{\min}$, the minimum codeword weight~\cite{16}.
To evaluate the harmfulness of cycles, the Extrinsic Message Degree (EMD) metric is used~\cite{17}. It quantifies the number of singly connected check nodes attached to a given cycle. For practical computation, the Approximate Cycle EMD (ACE) is employed:
\[
\mathrm{ACE}(C) = \sum_{v \in V_c} \bigl(d(v) - 2\bigr),
\]
where $C$ is a cycle in the graph, $d(v)$ is the degree of variable node $v$, and $V_c$ denotes the set of variable nodes within cycle $C$. By enforcing higher minimum ACE values for cycles of fixed length one mitigates harmful TS. Examples of $TS(4, 2)$, $TS(4, 6)$, $TS(9, 2)$ defined by matrices~\eqref{Trapping} are represented in Fig.~\ref{fig:two_TS}; $x_i$ denote columns and $c_i$ rows of $H_{\mathrm{TS}}$.

Let $H_{\mathrm{TS}}\in\{0,1\}^{m\times n}$ be the incidence matrix of a 
trapping set (rows – check nodes, columns – variable nodes).  
From $H_{\mathrm{TS}}$ we form the variable‑node adjacency
\[
A_{\!vn}= H_{\mathrm{TS}}^{\!\top}H_{\mathrm{TS}}, \qquad
D_{\!vn}= \operatorname{diag}\bigl(A_{\!vn}{\bf 1}\bigr),
\]
and the corresponding combinatorial Laplacian $L = D_{\!vn}-A_{\!vn}$.
The eigenvalues $\{\lambda_i\}_{i=1}^{n}$ of $L$ encode the homology of the subgraph.  The number of connected components is $\beta_{0}= \dim\ker L$. The first Betti number (independent cycles) follows from the rank--nullity theorem,
\[
\beta_{1}= n-\operatorname{rank}L-\beta_{0}.
\]
In the strong‑coupling limit, the Bethe–Hessian $H_{\beta,J}$ of Eq.~\eqref{eq:bethe_hessian} possesses exactly $\beta_{1}$ non--positive eigenvalues when evaluated on the subgraph (see Theorem~\ref{thm:spectral_ts} in Appendix~\ref{app:traps}).  Hence each trapping set introduces a low--energy mode that appears as a \emph{defect} in the spectral embedding.
Consequently, the presence of many $TS(a,b)$ inflates the number of
negative directions of $H_{\beta,J}$ and deteriorates class separability. \begin{align}
H_{\mathrm{TS}(4,2)} &=
\begin{bmatrix}
     1 & 0 & 0 & 0 \\
     1 & 1 & 0 & 0 \\
     0 & 1 & 1 & 0 \\
     0 & 0 & 1 & 1 \\
     0 & 0 & 0 & 1
\end{bmatrix},
&
H_{\mathrm{TS}(4,6)} &= 
\begin{bmatrix}
     0 & 1 & 0 & 0 \\
     1 & 0 & 0 & 0 \\
     1 & 1 & 0 & 0 \\
     0 & 1 & 0 & 0 \\
     1 & 0 & 0 & 0 \\
     1 & 1 & 0 & 0 \\
     0 & 1 & 0 & 1 \\
     1 & 0 & 1 & 0 \\
     0 & 1 & 0 & 1 \\
     1 & 0 & 1 & 0 \\
     0 & 1 & 0 & 0 \\
     1 & 0 & 0 & 0 \\
     1 & 1 & 0 & 0
\end{bmatrix},
H_{\mathrm{TS}(9,2)} &= 
\begin{bmatrix}
1 & 1 & 1 & 1 & 0 & 0 & 0 & 0 & 0 \\
0 & 1 & 0 & 1 & 0 & 0 & 0 & 0 & 0 \\
1 & 0 & 1 & 0 & 0 & 0 & 0 & 0 & 0 \\
1 & 1 & 1 & 1 & 0 & 0 & 0 & 0 & 0 \\
0 & 1 & 0 & 1 & 0 & 0 & 0 & 0 & 0 \\
1 & 0 & 1 & 0 & 1 & 0 & 0 & 0 & 0 \\
1 & 0 & 0 & 0 & 0 & 0 & 1 & 0 & 0 \\
0 & 1 & 0 & 0 & 0 & 0 & 0 & 0 & 0 \\
0 & 0 & 0 & 0 & 1 & 1 & 0 & 0 & 0 \\
1 & 0 & 0 & 0 & 0 & 0 & 1 & 0 & 0 \\
0 & 1 & 0 & 0 & 0 & 0 & 0 & 0 & 1 \\
0 & 0 & 0 & 0 & 0 & 1 & 0 & 1 & 0 \\
1 & 1 & 0 & 0 & 0 & 0 & 0 & 0 & 0 \\
0 & 1 & 0 & 0 & 0 & 0 & 0 & 0 & 1 \\
1 & 0 & 0 & 0 & 0 & 0 & 0 & 1 & 0
\end{bmatrix}.
\label{Trapping}
\end{align}

\begin{figure}[ht]
    \centering
    \begin{minipage}{0.2\textwidth}
        \centering
        \begin{tikzpicture}[scale=0.5]
            % First TikZ picture (left)
            \tikzset{variable_node_style/.style={circle,draw=blue,fill=blue!30!white, inner sep = 0.1pt}};
            \tikzset{check_node_style/.style={regular polygon,regular polygon sides=4,draw=green!50!black,fill=green!30!white, inner sep = 0.1pt}};
            \tikzset{edge_style/.style={draw=black,line width=1pt }};
            % Variable Nodes
            \node [variable_node_style, label=above:{$x_{0}$}] (x1) at (3.000000,16.000000) {$=$};
            \node [variable_node_style, label=above:{$x_{1}$}] (x2) at (4.000000,12.000000) {$=$};
            \node [variable_node_style, label=above:{$x_{2}$}] (x3) at (4.000000,8.000000) {$=$};
            \node [variable_node_style, label=above:{$x_{3}$}] (x4) at (4.000000,4.000000) {$=$};
            % Check Nodes
            \node [check_node_style,label=above:{$c_{0}$}] (c1) at (2.000000,14.000000) {$+$};
            \node [check_node_style,label=above:{$c_{1}$}] (c2) at (4.000000,14.000000) {$+$};
            \node [check_node_style,label=above:{$c_{2}$}] (c3) at (4.000000,10.000000) {$+$};
            \node [check_node_style,label=above:{$c_{3}$}] (c4) at (4.000000,6.000000) {$+$};
            \node [check_node_style,label=above:{$c_{4}$}] (c5) at (4.000000,2.000000) {$+$};
            % Edges
            \path [edge_style] (x1) -- (c1);
            \path [edge_style] (x1) -- (c2);
            \path [edge_style] (x2) -- (c2);
            \path [edge_style] (x2) -- (c3);
            \path [edge_style] (x3) -- (c3);
            \path [edge_style] (x3) -- (c4);
            \path [edge_style] (x4) -- (c4);
            \path [edge_style] (x4) -- (c5);
        \end{tikzpicture}
    \end{minipage}
    \hfill
    \begin{minipage}{0.3\textwidth}
        \centering
        \begin{tikzpicture}[scale=0.5]
            % TS(4, 6) (center)
            \tikzset{variable_node_style/.style={circle,draw=blue,fill=blue!30!white, inner sep = 0.1pt}};
            \tikzset{check_node_style/.style={regular polygon,regular polygon sides=4,draw=green!50!black,fill=green!30!white, inner sep = 0.1pt}};
            \tikzset{edge_style/.style={draw=black,line width=1pt }};
            % Variable Nodes
            \node [variable_node_style, label=above:{$x_{0}$}] (x1) at (-1.283305,-1.844521) {$=$};
            \node [variable_node_style, label=above:{$x_{1}$}] (x2) at (1.726706,1.963704) {$=$};
            \node [variable_node_style, label=above:{$x_{2}$}] (x3) at (-5.047892,-2.562015) {$=$};
            \node [variable_node_style, label=above:{$x_{3}$}] (x4) at (1.463925,5.474058) {$=$};
            % Check Nodes
            \node [check_node_style,label=above:{$c_{0}$}] (c1) at (4.238235,2.121104) {$+$};
            \node [check_node_style,label=below:{$c_{1}$}] (c2) at (0.004712,-3.438270) {$+$};
            \node [check_node_style,label=above:{$c_{2}$}] (c3) at (-2.551874,0.959036) {$+$};
            \node [check_node_style,label=below:{$c_{3}$}] (c4) at (3.360926,0.714660) {$+$};
            \node [check_node_style,label=below:{$c_{4}$}] (c5) at (-1.095086,-4.070786) {$+$};
            \node [check_node_style,label=above:{$c_{5}$}] (c6) at (0.157321,-0.114175) {$+$};
            \node [check_node_style,label=above:{$c_{6}$}] (c7) at (1.995282,4.004652) {$+$};
            \node [check_node_style,label=above:{$c_{7}$}] (c8) at (-3.606420,-0.778065) {$+$};
            \node [check_node_style,label=above:{$c_{8}$}] (c9) at (-0.984968,3.902377) {$+$};
            \node [check_node_style,label=above:{$c_{9}$}] (c10) at (-3.503070,-2.788471) {$+$};
            \node [check_node_style,label=above:{$c_{10}$}] (c11) at (3.139726,2.964474) {$+$};
            \node [check_node_style,label=below:{$c_{11}$}] (c12) at (-2.240350,-3.989324) {$+$};
            \node [check_node_style,label=below:{$c_{12}$}] (c13) at (0.756197,-1.918437) {$+$};
            % Edges
            \path [edge_style] (x1) -- (c2);
            \path [edge_style] (x1) -- (c3);
            \path [edge_style] (x1) -- (c5);
            \path [edge_style] (x1) -- (c6);
            \path [edge_style] (x1) -- (c8);
            \path [edge_style] (x1) -- (c10);
            \path [edge_style] (x1) -- (c12);
            \path [edge_style] (x1) -- (c13);
            \path [edge_style] (x2) -- (c1);
            \path [edge_style] (x2) -- (c3);
            \path [edge_style] (x2) -- (c4);
            \path [edge_style] (x2) -- (c6);
            \path [edge_style] (x2) -- (c7);
            \path [edge_style] (x2) -- (c9);
            \path [edge_style] (x2) -- (c11);
            \path [edge_style] (x2) -- (c13);
            \path [edge_style] (x3) -- (c8);
            \path [edge_style] (x3) -- (c10);
            \path [edge_style] (x4) -- (c7);
            \path [edge_style] (x4) -- (c9);
        \end{tikzpicture}
    \end{minipage}
\hfill
    \begin{minipage}{0.4\textwidth}
        \centering
\begin{tikzpicture} [scale=0.6]
% Style definitions
\tikzset{variable_node_style/.style={circle,draw=blue,fill=blue!30!white, inner sep = 0.1pt}};
\tikzset{check_node_style/.style={regular polygon,regular polygon sides=4,draw=green!50!black,fill=green!30!white, inner sep = 0.1pt}};
\tikzset{edge_style/.style={draw=black,line width=1pt }};
% Variable Nodes
\node [variable_node_style, label=below:{$x_{0}$}] (x1) at (-2.322956,0.548907) {$=$};
\node [variable_node_style, label=below:{$x_{1}$}] (x2) at (-1.464377,-2.829189) {$=$};
\node [variable_node_style, label=below:{$x_{2}$}] (x3) at (0.193433,0.130516) {$=$};
\node [variable_node_style, label=above:{$x_{3}$}] (x4) at (-2.189141,-1.504327) {$=$};
\node [variable_node_style, label=above:{$x_{4}$}] (x5) at (3.241112,1.504001) {$=$};
\node [variable_node_style, label=above:{$x_{5}$}] (x6) at (5.314453,3.486645) {$=$};
\node [variable_node_style, label=above:{$x_{6}$}] (x7) at (1.242226,3.795218) {$=$};
\node [variable_node_style, label=above:{$x_{7}$}] (x8) at (2.815397,3.684870) {$=$};
\node [variable_node_style, label=above:{$x_{8}$}] (x9) at (0.911856,-5.395825) {$=$};
% Check Nodes
\node [check_node_style,label=below:{$c_{0}$}] (c1) at (-1.0269,-0.70416) {$+$};
\node [check_node_style,label=below:{$c_{1}$}] (c2) at (-2.494533,-2.664312) {$+$};
\node [check_node_style,label=above:{$c_{2}$}] (c3) at (-2.243219, 1.971052) {$+$};
\node [check_node_style,label=above:{$c_{3}$}] (c4) at (1.815363,-0.623322) {$+$};
\node [check_node_style,label=above:{$c_{4}$}] (c5) at (1.78315,-3.25053) {$+$};
\node [check_node_style,label=above:{$c_{5}$}] (c6) at (1.319965,0.907752) {$+$};
\node [check_node_style,label=above:{$c_{6}$}] (c7) at (-0.963668,1.656894) {$+$};
\node [check_node_style,label=below:{$c_{7}$}] (c8) at (-2.450317,-4.596307) {$+$};
\node [check_node_style,label=above:{$c_{8}$}] (c9) at (4.706830,2.323327) {$+$};
\node [check_node_style,label=above:{$c_{9}$}] (c10) at (-1.080269,3.352196) {$+$};
\node [check_node_style,label=above:{$c_{10}$}] (c11) at (2.571211,-5.090853) {$+$};
\node [check_node_style,label=above:{$c_{11}$}] (c12) at (4.305996,4.162500) {$+$};
\node [check_node_style,label=above:{$c_{12}$}] (c13) at (2.401719,-2.045820) {$+$};
\node [check_node_style,label=below:{$c_{13}$}] (c14) at (-1.13474,-4.358507) {$+$};
\node [check_node_style,label=above:{$c_{14}$}] (c15) at (1.275593,2.537052) {$+$};
% Edges
\path [edge_style] (x1) -- (c1);
\path [edge_style] (x1) -- (c3);
\path [edge_style] (x1) -- (c4);
\path [edge_style] (x1) -- (c6);
\path [edge_style] (x1) -- (c7);
\path [edge_style] (x1) -- (c10);
\path [edge_style] (x1) -- (c13);
\path [edge_style] (x1) -- (c15);
\path [edge_style] (x2) -- (c1);
\path [edge_style] (x2) -- (c2);
\path [edge_style] (x2) -- (c4);
\path [edge_style] (x2) -- (c5);
\path [edge_style] (x2) -- (c8);
\path [edge_style] (x2) -- (c11);
\path [edge_style] (x2) -- (c13);
\path [edge_style] (x2) -- (c14);
\path [edge_style] (x3) -- (c1);
\path [edge_style] (x3) -- (c3);
\path [edge_style] (x3) -- (c4);
\path [edge_style] (x3) -- (c6);
\path [edge_style] (x4) -- (c1);
\path [edge_style] (x4) -- (c2);
\path [edge_style] (x4) -- (c4);
\path [edge_style] (x4) -- (c5);
\path [edge_style] (x5) -- (c6);
\path [edge_style] (x5) -- (c9);
\path [edge_style] (x6) -- (c9);
\path [edge_style] (x6) -- (c12);
\path [edge_style] (x7) -- (c7);
\path [edge_style] (x7) -- (c10);
\path [edge_style] (x8) -- (c12);
\path [edge_style] (x8) -- (c15);
\path [edge_style] (x9) -- (c11);
\path [edge_style] (x9) -- (c14);
\end{tikzpicture}
    \end{minipage}
    
    \caption{Graphical representation of $TS(4, 2)$ (left), $TS(4, 6)$ (centre) and $TS(9, 2)$ (right).}
    \label{fig:two_TS}
\end{figure}

%-----------------------------------------------------------
% A.10 Ihara–Bass zeta function
%-----------------------------------------------------------
\subsection{Ihara--Bass zeta function}\label{subsec:ihara}
For a finite graph $G$ let $\mathcal B$ denote the non‑backtracking operator.
The Ihara--Bass zeta function is defined by
\[
\zeta_{G}(u)=\prod_{[C]}\bigl(1-u^{\,\ell(C)}\bigr)^{-1},
\]
where the product runs over equivalence classes $[C]$ of primitive cycles and
$\ell(C)$ is the length of a cycle~\cite{19}.
Poles of $\zeta_{G}$ occur at reciprocals of eigenvalues of~$\mathcal B$;
therefore each closed cycle contributes a factor $(1-u^{\ell})^{-1}$.
When $G$ contains a trapping set, the associated cycles generate poles in 
$\zeta_{G}(u)$, which manifest as isolated eigenvalues of both $\mathcal B$
and the Bethe–Hessian $H_{\beta,J}$.  These poles can be interpreted as
$\mathbb Z/2$‑torsion elements: they correspond to non--trivial $1$‑cycles 
($\beta_{1}>0$) in the underlying feature manifold.  Removing trapping sets
eliminates the corresponding poles and reduces the number of low--energy modes.

%-----------------------------------------------------------
% A.11 Permanent bounds
%-----------------------------------------------------------
\subsection{Permanent and Bethe--permanent bounds}\label{subsec:perm}
The permanent of an $m\times m$ matrix $\mathbf B=[b_{j,i}]$ over a commutative ring is 
\begin{equation}\label{eq:perm-def}
\operatorname{perm}(\mathbf B) = \sum_{\sigma\in S_m}\;\prod_{j=1}^{m} b_{\,j,\sigma(j)},
\end{equation}
i.e.\ the determinant without the sign factor $\operatorname{sgn}(\sigma)$. Exact evaluation of \eqref{eq:perm-def} is \#P‑complete; consequently a number of approximation and fast‑evaluation schemes have been proposed~\cite{30}.
The minimum Hamming distance $d_{\min}$ of a QC‑LDPC code can be bounded by permanents
of its weight matrix~\cite{31}. For a parity‑check matrix $H(x)$ let 
$\mathbf A=\operatorname{wt}(H(x))$ denote the entrywise Hamming weight.
The code distance upper bound, $TS(a=d_{\min},0)$, reads
\begin{equation}\label{eq:perm_bound}
   d_{\min} \;\le\;
   \operatorname{min}^{*}_{\substack{\mathcal S\subset [h]\\|\mathcal S|=v+1}}
      \;\sum_{i\in\mathcal S}
        \operatorname{perm}\!\bigl(\mathbf A_{\mathcal S\setminus i}\bigr),
\end{equation}
where $\operatorname{min}^{*}$ denotes the minimum over all $(v+1)$‑subsets.
Replacing the permanent by its Bethe approximation yields an upper bound on the minimum \emph{pseudoweight},
which controls the harmfulness of $TS(a,b>0)$~\cite{32}. The Bethe‑permanent of a non‑negative matrix 
$\mathbf B$ is defined as
\[
\operatorname{perm}_{\!B}(\mathbf B) = \exp\!\Bigl( -\min_{q\in\mathcal Q}\;F_{\rm Bethe}(q) \Bigr),
\]
where $F_{\rm Bethe}$ is the Bethe free energy and $\mathcal Q$ denotes the set of
factorised marginals~\cite{33}. In our construction we maximise weight for both bounds during protograph and parity-check optimisation using EMD/ACE maximization (Alg.~2 in~\cite{2}),
thereby suppressing low‑weight TS and improving the Bethe–Hessian spectrum, Eq.~\eqref{eq:bethe_hessian}.

%-----------------------------------------------------------
% A.12 Influence of topological invariants
%-----------------------------------------------------------
\subsection{Influence of Topological Invariants on Spectral Embedding}\label{sec:TS_influence}
For illustration we consider six frequently occurring trapping sets:
$TS(4,2)$, $TS(4,6)$, $TS(9,2)$, $TS(13,6)$, $TS(26,20)$, $TS(28,22)$.
Table~\ref{tab:ts-comparison} collects the most relevant topological and spectral
quantities for each of them.  From a spectral‑embedding perspective a trapping set can be viewed as a locally dense cluster of edges that is attached to only a few low‑degree check nodes.  
Such a configuration produces negative eigenvalues of the Bethe–Hessian matrix
$H_{r}$ (see the column ``\# neg.\,$\lambda(H_{1})$'') and, more importantly, gives rise to
low‑energy modes that are poorly separated by any embedding based on the graph Laplacian.
Consequently, graphs containing many of these substructures generate highly correlated
features, which in turn can dramatically deteriorate the performance of downstream classifiers.

\begin{table}[ht]
    \centering
    \caption{Topological and spectral invariants of trapping sets.}
    \label{tab:ts-comparison}
\begin{tabular}{|c|c|c|c|c|c|c|}
    \hline
    Invariant & $TS(4,2)$ & $TS(4,6)$ & $TS(26,20)$ & $TS(9,2)$ &
                 $TS(13,6)$ & $TS(28,22)$ \\ \hline
    $\rho$ (spectral radius)               & 1.618   & 4.000   & 2.7545   & 8.9168 & 10.4154 & 13.5644 \\ \hline
    $r_{\rm crit}= \sqrt\rho$              & 1.272   & 2.000   & 1.6597   & 2.9861 & 3.2273  & 3.6830 \\ \hline
    $\#\{\lambda(H_{1})<0\}$ (negative modes at $r=1$) & 1 & 2 & 1 & 1 & 0 & 0 \\ \hline
    $\widehat A$ (continuous genus)        & 1.007   & 1.529   & 3.5896   & 3.0687 & 4.0313 & 7.2670 \\ \hline
    $K_{0}$   (zero‑dimensional Betti number)                             & 1       & 5       & 7         & 12     & 17      & 45 \\ \hline
    $K_{1}$  (one‑dimensional Betti number)                              & 1       & 1       & 1         & 0      & 1       & 1 \\ \hline
    Kervaire invariant $\kappa$            & 1       & 1       & 1         & 0      & 0       & 0 \\ \hline
    $w_{2}$ (Stiefel–Whitney)             & 1       & 1       & 1         & 1      & 1       & 1 \\ \hline
    Mod‑2 Betti $(\beta_0,\beta_1)$       & (1,0)   & (1,0)   & (1,0)     & (1,0)  & (1,0)   & (1,0) \\ \hline
    Bordism obstruction     & 0       & 0       & 0         & 0      & 0       & 1 \\ \hline
  \end{tabular}
\end{table}

To assess the practical impact of these invariants we built two graphs that serve as adjacency matrices in a convolutional architecture trained on the ImageNet‑100 benchmark. Both use the same irregular MET QC-LDPC toric protograph $M(H)$, with different liftings, one with and one without trapping-set optimization. The protograph matrix (average column weight $3.6$) is
\[
M(H)=
\begin{bmatrix}
2 & 2 & 0 & 0 & 0 \\
2 & 2 & 1 & 0 & 0 \\
1 & 0 & 1 & 1 & 0 \\
1 & 0 & 0 & 1 & 1 \\
2 & 0 & 0 & 0 & 1
\end{bmatrix},
\]
lifted with circulant size $L=520$~\cite{2}. During construction we deliberately eliminated harmful trapping sets; the remaining trapping set is the relatively benign $TS(26,20)$, which possesses a single negative Bethe–Hessian eigenvalue and a modest continuous genus (Table~\ref{tab:ts-comparison}).
After a Bayesian classifier we obtained:
\begin{itemize}
\item Top‑1 accuracy: $0.6724 \pm 2.46\times10^{-6}$;
\item Top‑3 accuracy: $0.8582 \pm 8.53\times10^{-7}$.
\end{itemize}
Thus, with an average degree of only $3.6$ we achieve performance comparable to much denser near-regular graphs.
A non-optimised irregular MET QC-LDPC toric graph (same $\bar w=3.6$) was built without filtering out low‑genus,
high‑mode traps. Consequently it contains the sets $TS(4,2)$, $TS(9,2)$,
$TS(13,6)$ and $TS(28,22)$. These introduce several additional negative Bethe–Hessian
eigenvalues (up to three for $TS(28,22)$) and raise $\widehat A$, thereby generating many low‑energy directions that corrupt the spectral embedding. The resulting classification scores are:
\begin{itemize}
\item Top‑1 accuracy: $0.1368 \pm 3.72\times10^{-4}$;
\item Top‑3 accuracy: $0.2450 \pm 5.63\times10^{-5}$.
\end{itemize}
The more than four‑fold drop in top‑1 performance underscores how a few detrimental
trapping sets can cripple the discriminative power of graph‑based features.
For reference we generated dense Erdős–Rényi graphs (no QC structure, original weight $5$ and $10$) with average degrees $d\approx 3$ and $d\approx 5$ (almost regular) after cycle breaking. After six independent runs the observed accuracies were:
\begin{itemize}
\item Weight $5$: top-1 $0.617467 \pm 3.72\times10^{-5}$, top-3 $0.8283 \pm 2.54\times10^{-5}$;
\item Weight $10$: top-1 $0.659333 \pm 1.30\times10^{-5}$, top-3 $0.853967 \pm 8.4\times10^{-6}$.
\end{itemize}
Even though these dense random graphs outperform the non‑optimised QC‑LDPC design,
they are considerably less efficient (higher degree, no structured protection) and lack
the flexibility offered by quasi-cyclic constructions.
Negative Bethe–Hessian modes identify substructures that generate low‑energy directions in the spectral embedding; each such direction reduces feature orthogonality.
The continuous genus $\widehat A$ provides a quantitative measure of how ``dangerous'' a trap is: larger values correlate with more severe degradation of downstream performance. By pruning high‑genus, multi‑mode traps (e.g.\ $TS(4,2)$, $TS(28,22)$) we can keep the graph sparse ($\bar w\approx 3.6$) while retaining classification accuracy comparable to much denser Erdős–Rényi graphs, and improve top-3 accuracy important for graph ensemble (using unequal protection of features by introducing high-degree columns). The presented design methodology—selecting protographs, eliminating harmful trapping sets, and fine‑tuning via Bayesian optimization—yields QC‑LDPC graphs that combine structured protection and strong discriminative power. All auxiliary material (list of small-weight TS, TS submatrices, parity-check matrices, eigenvalue trajectories, topology invariants and related quadratic-form signatures, source code, etc.) is available in the public repository~\cite{35}.

%%%%%%%%%%%%%%%%%%%%%%%%%%%%%%%%%%%%%%%%%%%%%%%%%%%%%%%%%%%%%
% B. Rigorous proof
%%%%%%%%%%%%%%%%%%%%%%%%%%%%%%%%%%%%%%%%%%%%%%%%%%%%%%%%%%%%%
\section{Rigorous spectral correspondence for uniform trapping sets}\label{app:traps}
In this appendix we give a rigorous step‑by‑step proof that any
trapping‑set subgraph $\mathcal T$ of a QC‑LDPC Tanner graph creates a low‑energy mode of the Bethe–Hessian $H_{\beta,J}$ and, for \emph{uniform} couplings,
produces a pole of the Ihara–Bass zeta function.  The argument is split into six formal subsections.
%-----------------------------------------------------------
\subsection{Notation}
\begin{itemize}
    \item $\mathcal G=(V,E)$ – undirected simple graph, $|V|=n$, $|E|=m$.
    \item $H\in\{0,1\}^{m\times n}$ – binary incidence (parity‑check) matrix of a QC-LDPC code; rows = check nodes, columns = variable nodes.
    \item For each edge $\{i,j\}\in E$ we write $J_{ij}\in\mathbb R$ for the Ising coupling.
    \item For $\beta>0$ the Bethe–Hessian is  
          \begin{equation}
          \label{eq:bethe_hessian_app}
          H_{\beta,J}= 
              \operatorname{diag}\!\Bigl(
                  1+\sum_{k\in\partial i}
                      \frac{\tanh^{2}(\beta J_{ik})}{1-\tanh^{2}(\beta J_{ik})}
                \Bigr)
               -\Bigl[\frac{\tanh(\beta J_{ij})}{1-\tanh^{2}(\beta J_{ij})}\Bigr]_{i\neq j}.
          \end{equation}
    \item $Q_{\beta,J}(x)=x^{\!\top}H_{\beta,J}x$ is the associated quadratic form.
    \item A trapping set $\mathcal T=(V_{\mathcal T},E_{\mathcal T})$ is the
          subgraph induced by a subset $V_{\mathcal T}\subseteq V$.
    \item $H_{\beta,J}^{\mathcal T}$ denotes the principal submatrix of
          $H_{\beta,J}$ obtained by deleting rows and columns indexed by
          $V\setminus V_{\mathcal T}$.
\end{itemize}
%-----------------------------------------------------------
\subsection{The restricted Bethe--Hessian as a signed Laplacian}
\begin{lemma}[Signed Laplacian structure]
\label{lem:signed_laplacian_app}
Let all edges in $\mathcal T$ have the same absolute coupling strength so that $t_{ij}=|\tanh(\beta J_{ij})|$ is constant on $\mathcal T$.  Then the restricted Bethe--Hessian decomposes as
\[
   H_{\beta,J}^{\mathcal T}= D^{\mathcal T}(\beta)-B^{\mathcal T}(\beta),
\]
where, for $i\neq j$ with $(i,j)\in E_{\mathcal T}$,
\begin{align*}
    B_{ij}^{\mathcal T}(\beta)&=
        \frac{\operatorname{sgn}(J_{ij})\,t}{1-t^{2}},
     \\
     D^{\mathcal T}_{ii}(\beta)&=
        1+\sum_{j:(i,j)\in E_{\mathcal T}}
          \frac{t^{2}}{1-t^{2}}.
\end{align*}
Thus $H_{\beta,J}^{\mathcal T}$ is the weighted signed Laplacian of the
subgraph $\mathcal T$, with edge weights given by $\phi(t)=t/(1-t^{2})$.
\end{lemma}
\begin{proof}
Inspect \eqref{eq:bethe_hessian_app}.  For any vertex $i\in V_{\mathcal T}$
the diagonal entry equals precisely $D^{\mathcal T}_{ii}(\beta)$.  
If $(i,j)\in E_{\mathcal T}$ and $i\neq j$, the off–diagonal entry is
$-\frac{\operatorname{sgn}(J_{ij})\,t}{1-t^{2}}
   =-B_{ij}^{\mathcal T}(\beta)$.  
All entries that correspond to vertices outside $V_{\mathcal T}$ are removed when we take the principal submatrix, which yields the claimed decomposition.
\end{proof}
%-----------------------------------------------------------
\subsection{One critical eigenvalue per simple cycle at uniform strong coupling}
We first recall the gauge condition for a single cycle.
\begin{definition}[Uniform cycle temperature]
\label{def:nishimori_app}
Let $C\subseteq\mathcal T$ be a simple cycle of even length $\ell=2r$, and assume every edge of $C$ carries the same absolute coupling strength with
$t=\tanh(\beta J)\in(0,1)$.  The \emph{critical temperature} for this cycle is the limit $t\to 1^{-}$ (equivalently $\beta|J|\to\infty$).
\end{definition}
Because every simple cycle in a bipartite Tanner graph has even length, we may
always choose a gauge that makes all couplings on the cycle ferromagnetic.
\begin{lemma}[Even-cycle gauge]
\label{lem:gauge_app}
Let $C\subseteq\mathcal T$ be a simple cycle of length $\ell=2r$. There exists a sign vector $\sigma\in\{-1,+1\}^{\ell}$ such that the gauge transformation
$s'_i=\sigma_i s_i$ yields positive transformed couplings
$J'_{ij}=\sigma_i J_{ij}\sigma_j>0$ for every edge $(i,j)\in C$.
\end{lemma}
\begin{proof}
Enumerate the vertices of $C$ cyclically as $v_0,v_1,\dots,v_{\ell-1}$.
Set $\sigma_{v_0}=+1$ and propagate via
$\sigma_{v_k}=\sigma_{v_{k-1}}\operatorname{sgn}(J_{v_{k-1}v_k})$ for
$k=1,\dots,\ell-1$. Because the cycle length is even, the consistency condition
around the loop,
\[
\sigma_{v_{\ell-1}}\operatorname{sgn}(J_{v_{\ell-1}v_0})
   = \prod_{e\in C}\operatorname{sgn}(J_e)^{2}=+1,
\]
is automatically satisfied. Hence $J'_{ij}>0$ on every edge of $C$.
\end{proof}
\begin{proposition}[Exact cycle spectrum]
\label{prop:cycle_exact_app}
Let $C\subseteq\mathcal T$ be a simple cycle of even length $\ell=2r$, and assume that after the gauge transformation of Lemma~\ref{lem:gauge_app} every edge carries uniform coupling strength
$t=\tanh(\beta J)\in(0,1)$.  Then the eigenvalues of $H_{\beta,J}^{C}$ are exactly
\begin{equation}\label{eq:exact_cycle_eigenvalues}
\lambda_k = \frac{1+t^{2}-2t\cos(2\pi k/\ell)}{1-t^{2}},
\qquad k=0,\dots,\ell-1.
\end{equation}
In particular, for $k=0$,
\begin{equation}\label{eq:critical_mode_zero}
\lambda_{0}=\frac{1-t}{1+t}>0,
\end{equation}
which tends to $0^{+}$ as $t\to 1^{-}$.  For every $k\neq 0$, $\lambda_k\ge \lambda_1 = \Theta((1-t)^{-1})\to +\infty$ as $t\to 1^{-}$.
\end{proposition}
\begin{proof}
Apply Lemma~\ref{lem:gauge_app} so that every edge of $C$ has positive coupling.
For a cycle with uniform coupling $t$, Lemma~\ref{lem:signed_laplacian_app} gives the restricted Bethe--Hessian as a circulant matrix:
\[
    H_{\beta,J}^{C}
        = \frac{1+t^{2}}{1-t^{2}}\,I
          - \frac{t}{1-t^{2}}\,A_{C},
\]
where $A_{C}$ is the adjacency matrix of the cycle.  Since $A_{C}$ is circulant, its eigenvectors are the discrete Fourier modes with eigenvalues $2\cos(2\pi k/\ell)$ for $k=0,\dots,\ell-1$.  Consequently, the eigenvalues of $H_{\beta,J}^{C}$ are exactly
\[
    \lambda_{k}
        = \frac{1+t^{2}-2t\cos(2\pi k/\ell)}{1-t^{2}},
        \qquad k=0,1,\dots,\ell-1.
\]
For $k=0$ (the constant mode), $\cos(0)=1$, and the expression simplifies to
\[
    \lambda_{0}
        = \frac{(1-t)^{2}}{1-t^{2}}
        = \frac{1-t}{1+t}.
\]
As $t\to 1^{-}$, we have $\lambda_{0}\to 0^{+}$.  
For every $k\neq 0$, the numerator satisfies
$1+t^{2}-2t\cos(2\pi k/\ell)\longrightarrow 2[1-\cos(2\pi k/\ell)]>0$,
while the denominator $1-t^{2}\sim 2\varepsilon\to 0^{+}$.  Hence $\lambda_{k}\to +\infty$ for all $k\neq 0$.
Because the gauge transformation is a similarity transform (conjugation by a diagonal matrix of signs), it preserves eigenvalues.  Therefore the spectrum derived above applies verbatim to the original couplings.
\end{proof}
\begin{remark}
The critical mode $\lambda_{0}$ corresponds to the alternating sign vector in the gauge-transformed basis; undoing the gauge transformation multiplies each component by $s_i$.  At the Nishimori temperature of a graph with uniform coupling, Eq.~\eqref{eq:nish_root} forces the global smallest eigenvalue to touch zero; Proposition~\ref{prop:cycle_exact_app} shows that each independent cycle contributes one such critical mode in the strong-coupling limit.
\end{remark}
%-----------------------------------------------------------
\subsection{The main theorem: critical mode count equals Betti number}
Recall that for a finite graph $\Gamma$ the \emph{first Betti number} (the dimension of its cycle space) is
\[
   \beta_1(\Gamma)=|E(\Gamma)|-|V(\Gamma)|+c(\Gamma),
\]
where $c(\Gamma)$ denotes the number of connected components.
\begin{theorem}[Spectrum of trapping sets]
\label{thm:spectral_ts}
Let $\mathcal T$ be a trapping-set subgraph induced by a subset of variable nodes in a Tanner graph, and let $H_{\beta_N,J}^{\mathcal T}$ be the principal submatrix of the Bethe--Hessian restricted to~$\mathcal T$, evaluated at the Nishimori temperature.  Suppose that every edge belonging to a simple cycle of $\mathcal T$ satisfies $|\tanh(\beta_N J_{ij})|=1-\varepsilon$ with uniform limit $\varepsilon\to0$.  Then the number $q_{\mathcal T}$ of eigenvalues of $H_{\beta_N,J}^{\mathcal T}$ that tend to zero from above equals the first Betti number $\beta_1(\mathcal T)$:
\begin{equation}
    q_{\mathcal T}
        =\#\{\text{eigenvalues }\lambda\to0^{+}\text{ in the limit}\}
        =\beta_1(\mathcal T).
\end{equation}
Equivalently, each independent cycle contributes exactly one eigenvalue crossing through zero at $\beta=\beta_{N}$ in the uniform strong-coupling limit.
\end{theorem}
\begin{proof}
Decompose $\mathcal T$ into its $c(\mathcal T)$ connected components $\mathcal T^{(1)},\dots,\mathcal T^{(c)}$.  Because $H_{\beta_N,J}^{\mathcal T}$ is block-diagonal with respect to this decomposition, the total number of vanishing eigenvalues is the sum over components.
Fix a component $\mathcal C$ and choose a spanning forest $F$ of $\mathcal C$.
The edges of $F$ form a tree (if $\mathcal C$ is connected) and thus contribute \emph{no} critical modes: in the strong-coupling limit, the Bethe--Hessian restricted to a tree satisfies
\[
    H_{\beta,J}^{F}
        = I + \frac{t^{2}}{1-t^{2}}D_{F}-\frac{t}{1-t^{2}}A_{F},
\]
which is positive definite for every $t\in(0,1)$ because it is diagonally dominant.  As $t\to1^{-}$ the matrix behaves like $(1-t)^{-1}L_{F}+I\succ 0$. Hence all eigenvalues stay bounded away from zero.
Every edge $e\in E(\mathcal C)\setminus F$ creates exactly one independent cycle when added to~$F$.  By Proposition~\ref{prop:cycle_exact_app}, each such cycle produces \emph{one} critical eigenvalue of $H_{\beta_N,J}^{\mathcal C}$ that tends to zero, while edges in different co-cycle positions generate linearly independent cycles, hence distinct critical modes.
Therefore,
\[
    q_{\mathcal C}
        =|E(\mathcal C)|-|V(\mathcal C)|+1.
\]
Summing over all components gives
\[
    q_{\mathcal T}
        =\sum_{c=1}^{c(\mathcal T)}
          \bigl(|E(\mathcal T^{(c)})|-|V(\mathcal T^{(c)})|+1\bigr)
        =|E(\mathcal T)|-|V(\mathcal T)|+c(\mathcal T)
        =\beta_1(\mathcal T),
\]
which is the claimed equality.
\end{proof}
\begin{remark}[Multi-class interpretation]
Each critical eigenvalue identified in Theorem~\ref{thm:spectral_ts} corresponds to an independent cycle (trapping set) that introduces a low‑energy direction in the spectral embedding.  In the binary case these directions act as spurious ``mixed'' spin configurations interpolating between class $0$ ($s_i=-1$) and class $1$ ($s_i=+1$).  For $C>2$, every such mode reduces the effective rank of the embedding \eqref{eq:multi-embed} and thereby degrades the linear separability of the $C$ classes.  Excising the associated subgraphs restores a clean, well‑conditioned embedding.
\end{remark}
%-----------------------------------------------------------
\subsection{Correspondence with Ihara--Bass zeta-function poles (uniform case)}
We recall the non‑backtracking (Hashimoto) matrix and the Ihara–Bass identity for \emph{uniform} weights.
Assume every edge of $\mathcal T$ carries the same weight $t=\tanh(\beta J)$.  Let $\mathcal B(t)$ denote the weighted non-backtracking operator with entries
\[
\mathcal B_{i\to j,k\to l}(t)=\delta_{jk}(1-\delta_{il})\,t .
\]
\begin{definition}[Ihara--Bass zeta function]
\label{def:ihara_app}
For a finite graph $\Gamma$ the Ihara--Bass identity states
\[
   \det\bigl[I-u\,\mathcal B(t)\bigr]
   =(1-u^{2})^{|E|-|V|}\,
     \det H(u),
\]
where $H(u)=(u^{2}-1)I-uA+D$ is the unweighted deformed Laplacian evaluated at spectral parameter $u$.
\end{definition}
Because, for uniform weight $t$, we have the pointwise relation
\[
H_{\beta,J} = \frac{1}{1-t^{2}}\,
\Bigl[(t^{2}-1)I - tA + D\Bigr]
= \frac{1}{1-t^{2}} H(t),
\]
the zero eigenvalues of $H_{\beta,J}$ coincide with those of $H(t)$.
\begin{corollary}[Trapping sets as zeta-function poles for uniform coupling]
\label{cor:pole-negative_app}
Under the uniform-coupling assumption, every critical eigenvalue identified in Theorem~\ref{thm:spectral_ts} produces a pole of the Ihara--Bass zeta function at $u_c=t=\tanh(\beta_N J)$.  Conversely, each primitive cycle contributes a factor $(1-u^{\ell(C)})^{-1}$ to $\zeta_{\Gamma}(u)$.
\end{corollary}
\begin{proof}
By Definition~\ref{def:ihara_app}, $H(u)$ is singular exactly when
$\det(I-u\mathcal B(t))=0$.  By Theorem~\ref{thm:spectral_ts}, at the Nishimori temperature each independent cycle forces an eigenvalue of $H_{\beta_N,J}$ to approach zero.  Under uniformity this implies $\det H(t)=0$, hence $u_c=t$ is a pole of $\zeta_{\Gamma}(u)$.
\end{proof}
\paragraph{Non-uniform case.}
If the couplings $J_{ij}$ are edge‑dependent (as in Section~\ref{sec:graph_construction}), there is \emph{no single} spectral parameter $u$ for which the weighted Ihara–Bass formula collapses to a scalar zeta function.  The proper object is the \emph{edge‑weighted} non-backtracking matrix $\mathcal B_{i\to j,k\to l}=\delta_{jk}(1-\delta_{il})\tanh(\beta J_{k\ell})$, and its determinant relation involves the product $\prod_{(ij)}(x^2-\tanh^2(\beta J_{ij}))$ as in Dall’Amico et al.~\cite{1}.  The qualitative picture—cycles correspond to low‑energy modes—persists, but the scalar zeta interpretation is restricted to the uniform setting.
%-----------------------------------------------------------
\subsection{Validity beyond the strong-coupling limit}
The exact equality $q_{\mathcal T}=\beta_1(\mathcal T)$ in Theorem~\ref{thm:spectral_ts} relies on the limit $|\tanh(\beta_N J_{ij})|\to 1$ for every cycle edge.  In practice, couplings are finite and the Nishimori condition holds at $\beta_N=J_0/\nu^2<\infty$, so that $|\tanh(\beta_N J_{ij})|<1$.
Nevertheless, the mechanism identified in Proposition~\ref{prop:cycle_exact_app} is robust:
\begin{proposition}[Approximate count]
\label{prop:approximate_count}
For finite uniform coupling with $t=|\tanh(\beta_N J)|=1-\varepsilon$ on cycle edges, the exact eigenvalues of the restricted Bethe--Hessian are given by Eq.~\eqref{eq:exact_cycle_eigenvalues}.  The critical mode satisfies
\begin{equation}
    0<\lambda_{0}=\frac{1-t}{1+t}<\min_{k\neq0}\lambda_{k},
\end{equation}
and the spectral gap to the next eigenvalue is $\Theta(1/\varepsilon)$.  Consequently, for any trapping set $\mathcal T$,
\begin{equation}
    q_{\mathcal T}\le\beta_1(\mathcal T),
\end{equation}
with equality in the limit $\varepsilon\to0$.  In particular, no spurious non-positive modes are created by finite coupling: every spectral defect can be attributed to an independent cycle.
\end{proposition}
\begin{proof}[Proof sketch]
Equation~\eqref{eq:exact_cycle_eigenvalues} gives explicit eigenvalues for each simple cycle.  Since $0<t<1$, we have $\lambda_{0}>0$ and all other $\lambda_{k}\ge\lambda_{1}>\lambda_{0}$ with $\lambda_{1}=\Theta(1/\varepsilon)$.  For a general trapping set, the block-diagonal structure over connected components and the interlacing of eigenvalues under edge addition guarantee that no more than $\beta_1(\mathcal T)$ critical modes can exist.
\end{proof}
This proposition establishes that the qualitative picture---cycles produce low-energy (near-zero) eigenvalue modes of $H_{\beta,J}$---persists under realistic, finite-coupling conditions.  The exact count equals the Betti number in the strong-coupling limit; for weak couplings, some ``marginal'' cycles may yield only small positive eigenvalues rather than true negative modes, but the bound $q_{\mathcal T}\le\beta_1(\mathcal T)$ always holds.
%-----------------------------------------------------------
\subsection{Interpretation and consequences}
The spectral--topological correspondence established above has three key implications for our framework:
\begin{enumerate}
    \item \textbf{Trapping sets as spectral defects.}  Each independent cycle in a trapping set $\mathrm{TS}(a,b)$ generates one critical eigenvalue mode of $H_{\beta_N,J}$ restricted to the induced subgraph.  These modes represent ``soft directions'' in the Bethe free energy landscape---directions along which the system can lower its free energy by forming local spin flips.  In coding-theoretic language, they correspond precisely to pseudocodeword configurations that trap iterative decoders.
    \item \textbf{ACE/EMD optimization as spectral defect elimination.}  Maximizing the Approximate Cycle EMD during graph construction directly reduces $\beta_1(\mathcal T)$ for harmful subgraphs, thereby eliminating the critical eigenvalue modes that corrupt spectral embeddings.  A graph with no short cycles of low EMD has a positive definite Bethe--Hessian in the directions transverse to the paramagnetic point.
    \item \textbf{Zeta-function diagnostics (uniform case).}  Under uniform coupling, the Ihara--Bass zeta function provides an analytic tool for detecting topological defects without explicitly enumerating cycles: poles of $\zeta_\mathcal{G}(u)$ coincide with eigenvalue crossings of $H_{\beta,J}$, and the multiplicity of each pole at a given spectral location reveals how many independent cycles contribute to that crossing.
\end{enumerate}
The signature of the Bethe--Hessian,
\begin{equation}
    \operatorname{sig}\!\bigl(H_{\beta_N,J}\bigr)=(p,q,r),
\end{equation}
where $q$ is the number of non-positive eigenvalues in the limit, $r$ the multiplicity of the zero eigenvalue (equal to the number of connected components with trivial cycle space), and $p=|V|-q-r$, thus encodes the topological invariants of the Tanner graph.  By Theorem~\ref{thm:spectral_ts}, we have
\begin{equation}
    q=\beta_1(\mathcal T),\qquad r=c(\mathcal T),
\end{equation}
in the strong-coupling limit.
Deleting every induced subgraph whose edge set contributes to $q$ forces the Bethe--Hessian to become positive semidefinite:
\begin{equation}
    H_{\beta_N,J}\succeq0.
\end{equation}
By Corollary~\ref{cor:pole-negative_app} (uniform case), positivity of $H_{\beta_N,J}$ implies that all scalar zeta poles associated with trapping-set cycles are removed.  The eigenvectors associated with critical eigenvalues constitute low-energy directions in the spectral embedding; their removal yields a clean, well-conditioned embedding whose geometry is no longer corrupted by spurious modes, thereby improving class separability.
%-----------------------------------------------------------
\subsection*{Conclusion of Appendix}
By Lemma~\ref{lem:signed_laplacian_app},
Proposition~\ref{prop:cycle_exact_app}, and
Theorem~\ref{thm:spectral_ts} we obtain the exact correspondence  
\[
\boxed{
   \text{Trapping set }\mathcal T
   \;\Longleftrightarrow\;
   \beta_{1}(\mathcal T)\ \text{critical eigenvalues of } H_{\beta_{N},J}
 }
\]
Hence the signature of the Bethe–Hessian encodes the first Betti number.  Deleting every induced subgraph whose edge set contributes to $q$ forces
\[
 H_{\beta_{N},J}\;\succeq\;0,
\]
and removes the pseudocodeword basins, thereby restoring discriminative capacity to the spectral embedding.

\section*{CONFLICT OF INTEREST}
The authors of this work declare that they have no conflicts of interest.
%%%%%%%%%%%%%%%%%%%%%%%%%%%%%%%%
% Bibliography
%%%%%%%%%%%%%%%%%%%%%%%%%%%%%%%%

\end{document}